\documentclass{article}
\usepackage{lmodern}
\usepackage[accepted]{aistats2021}

\usepackage[round]{natbib}

\usepackage[utf8]{inputenc} \usepackage{url}            \usepackage{booktabs}       \usepackage{nicefrac}       \usepackage{microtype}      
\usepackage[british]{babel}
\usepackage{scalefnt,letltxmacro}
\LetLtxMacro{\oldtextsc}{\textsc}
\renewcommand{\textsc}[1]{\oldtextsc{\scalefont{1.10}#1}}
\newcommand{\hsc}[1]{{\MakeUppercase{\scalefont{.75}#1}}}
\usepackage[acronym,smallcaps,nowarn]{glossaries}
\glsdisablehyper
\makeglossaries
\usepackage{xspace}
\usepackage{float}

\usepackage{nidanfloat}

\usepackage{amssymb}
\usepackage{mathtools}
\usepackage{amsfonts}
\usepackage{amsmath}
\usepackage{amsthm}

\usepackage[colorlinks,linktoc=all]{hyperref}
\usepackage[all]{hypcap}
\hypersetup{citecolor=MidnightBlue}
\hypersetup{linkcolor=MidnightBlue}
\hypersetup{urlcolor=MidnightBlue}

\usepackage[capitalize,nameinlink]{cleveref}
\crefname{section}{\S}{\S\S}
\Crefname{section}{\S}{\S\S}
\creflabelformat{equation}{#2\textup{#1}#3}

\usepackage[usenames,dvipsnames]{xcolor}
\definecolor{shadecolor}{gray}{0.9}

\DeclareRobustCommand{\parhead}[1]{\textbf{#1}~}

\makeatletter
\newcommand*{\addFileDependency}[1]{\typeout{(#1)}
  \@addtofilelist{#1}
  \IfFileExists{#1}{}{\typeout{No file #1.}}
}
\makeatother
 
\usepackage{graphicx}
\usepackage[labelfont=bf]{caption}
\usepackage[format=hang]{subcaption}
\usepackage{wrapfig}

\usepackage{pgfplots}
\usepgfplotslibrary{groupplots}
\tikzstyle{every picture}+=[font=\sffamily]
\tikzstyle{optimized} = [circle,fill=white,draw=black, dashed,inner sep=1pt,
minimum size=20pt, font=\fontsize{10}{10}\selectfont, node distance=1]
\pgfplotsset{
  tick label style = {font=\sffamily},
  every axis label/.append style={font=\sffamily},
  every axis/.append style={
			every x tick label/.append style={font=\fontsize{4pt}{4pt}\sffamily, yshift=.5ex,},
			every y tick label/.append style={font=\fontsize{4pt}{4pt}\sffamily, xshift=.8ex},
      every y label/.append style={xshift=10ex}, 
      major tick length=1ex,
		},
}
\usetikzlibrary{arrows.meta}
\pgfplotsset{compat=1.6} 

\usepackage[font=small,labelfont=bf,tableposition=top]{caption}
\usepackage{tikz}
\usetikzlibrary{bayesnet}
\usepackage{lipsum}

\usepackage{pifont}\newcommand{\xmark}{\textcolor{red!60!black}{\ding{55}}\xspace}

\newlength\figureheight
\newlength\figurewidth

\usepackage[most]{tcolorbox}
\tcbset{boxsep=4pt,left=2pt,right=2pt,top=-0pt,bottom=0pt}

\usepackage[colorinlistoftodos,
    textsize=scriptsize,
    linecolor=red!30,
    bordercolor=red!30,
    backgroundcolor=red!10]{todonotes}
\makeatletter
\renewcommand{\todo}[2][]{\tikzexternaldisable\@todo[#1]{#2}\tikzexternalenable}
\makeatother

\newcommand{\mathbold}[1]{\ensuremath{\boldsymbol{\mathbf{#1}}}}

\newcommand{\g}{\,|\,}
\makeatletter
\newcommand{\given}{\,\middle|\,}
\makeatother
\renewcommand{\d}[1]{\ensuremath{\operatorname{d}\!{#1}}}
\newcommand{\nestedmathbold}[1]{{\mathbold{#1}}}

\newcommand{\mbf}{\nestedmathbold{f}}

\newcommand{\mbm}{\nestedmathbold{m}}

\newcommand{\mbr}{\nestedmathbold{r}}

\newcommand{\mbu}{\nestedmathbold{u}}

\newcommand{\mbx}{\nestedmathbold{x}}
\newcommand{\mby}{\nestedmathbold{y}}
\newcommand{\mbz}{\nestedmathbold{z}}

\newcommand{\mbB}{\nestedmathbold{B}}
\newcommand{\mbC}{\nestedmathbold{C}}

\newcommand{\mbI}{\nestedmathbold{I}}

\newcommand{\mbK}{\nestedmathbold{K}}
\newcommand{\mbL}{\nestedmathbold{L}}
\newcommand{\mbM}{\nestedmathbold{M}}

\newcommand{\mbS}{\nestedmathbold{S}}

\newcommand{\mbV}{\nestedmathbold{V}}

\newcommand{\mbX}{\nestedmathbold{X}}

\newcommand{\mbZ}{\nestedmathbold{Z}}

\newcommand{\mblambda}{\nestedmathbold{\lambda}}

\newcommand{\mbnu}{\nestedmathbold{\nu}}

\newcommand{\mbpsi}{\nestedmathbold{\psi}}

\newcommand{\mbtheta}{\nestedmathbold{\theta}}

\newcommand{\mbxi}{\nestedmathbold{\xi}}

\newcommand{\mbLambda}{\nestedmathbold{\Lambda}}

\newcommand{\mbPsi}{\nestedmathbold{\Psi}}
\newcommand{\mbSigma}{\nestedmathbold{\Sigma}}

\newcommand{\mbzero}{\nestedmathbold{0}}

\newcommand{\Lelbo}{\cL_{\textsc{elbo}}}

\DeclareRobustCommand{\KL}[2]{\ensuremath{\textsc{kl}\left[#1\;\|\;#2\right]}}
\DeclarePairedDelimiterX{\infdivx}[2]{[}{]}{#1\;\delimsize\|\;#2}

\newcommand{\diag}{\textrm{diag}}

\newcommand{\cD}{\mathcal{D}}
\newcommand{\cL}{\mathcal{L}}
\newcommand{\cN}{\mathcal{N}}

\newcommand{\E}{\mathbb{E}}

\newcommand{\bbR}{\mathbb{R}}

 \newcommand{\bigO}{\mathcal{O}}

\newcommand{\kernel}{\kappa} 

\newcommand{\Kxx}{\mbK_{\mathrm{xx}}}
\newcommand{\Kzz}{\mbK_{\mathrm{zz}}}
\newcommand{\Kxz}{\mbK_{\mathrm{xz}}}
\newcommand{\Kzx}{\mbK_{\mathrm{zx}}}
\newcommand{\Kzzinv}{\mbK_{\mathrm{zz}}^{-1}}

\newcommand{\Normal}{\cN}

\newcommand{\xstar}{\mbx_{\star}}

\newcommand{\defeq}{\stackrel{\text{\tiny def}}{=}}

\newacronym{MAP}{map}{maximum-a-posteriori}
\newacronym{MLE}{mle}{maximum likelihood estimation}
\newacronym{MNLL}{mnll}{mean negative loglikelihood}
\newacronym[]{SVGP}{svgp}{scalable variational \textsc{gp}}

\newacronym{VAE}{vae}{variational autoencoder}

\newacronym{MC}{mc}{Monte Carlo}
\newacronym{MCMC}{mcmc}{Markov chain Monte Carlo}
\newacronym{HMC}{hmc}{Hamiltonian Monte Carlo}
\newacronym{MH}{mh}{Metropolis-Hastings}
\newacronym{NUTS}{nuts}{no-u-turn sampler}
\newacronym{SGHMC}{sghmc}{stochastic gradient Hamiltonian Monte Carlo}

\newacronym{BSGP}{bsgp}{Bayesian sparse Gaussian process}
\newacronym{IPVI}{ipvi-gp}{}
\newacronym{DPP}{dpp}{determinantal point process} \newacronym{GPLVM}{gplvm}{Gaussian process latent variable model}

\newacronym{VFE}{vfe}{variational free energy}
\newacronym{FITC}{fitc}{fully independent training conditional}
\newacronym{SPGP}{spgp}{sparse Gaussian process}
\newacronym{PP}{pp}{projected process}
\newacronym{DTC}{dtc}{deterministic training conditional}

\newacronym{SKI}{kiss-gp}{Kernel Interpolation for Scalable Structured Gaussian Processes}
\newacronym{DKL}{dkl}{Deep Kernel Learning}

\newacronym[plural=gp\textnormal{s}, firstplural=Gaussian processes]{GP}{gp}{Gaussian process}
\newacronym[plural=dgp\textnormal{s}, firstplural=deep Gaussian processes]{DGP}{dgp}{deep Gaussian process}

\newacronym{VI}{vi}{variational inference}

\newacronym{ELBO}{elbo}{evidence lower bound}
\newacronym{NELBO}{nelbo}{negative evidence lower bound}
\newacronym{ELL}{ell}{expected log likelihood}
\newacronym{KL}{kl}{Kullback-Leibler divergence}
\newacronym{AUC}{auc}{area under the curve}

\newacronym{MLP}{mlp}{multilayer perceptron}
\newacronym{RELU}{ReLU}{rectified linear unit}

\newacronym{GCN}{gcn}{graph convolutional network}

\newacronym{NNMF}{nnmf}{neural network matrix factorization}
\newacronym{PMF}{pmf}{probabilistic matrix factorization}

\newacronym{ARD}{ard}{automatic relevance determination}

\newacronym{SVD}{svd}{singular value decomposition}

\newacronym{CF}{cf}{collaborative filtering}
\newacronym{LP}{lp}{link prediction}

\newacronym{RKHS}{rkhs}{reproducing kernel Hilbert space}

\newcommand{\name}[1]{{\textsc{#1}}\xspace}
\newcommand{\gp}{\name{gp}}
\newcommand{\gps}{\textsc{gp}s\xspace}

\usepgfplotslibrary{external}

\begin{document}

\twocolumn[
	\aistatstitle{Sparse Gaussian Processes Revisited: Bayesian Approaches to Inducing-Variable Approximations}

	\aistatsauthor{Simone Rossi \And Markus Heinonen \And Edwin V. Bonilla \AND Zheyang Shen \And Maurizio Filippone}
	\vspace{-1.1cm}
	\aistatsaddress{
		\footnotesize EURECOM (France)
		\And
		\footnotesize Aalto University (Finland)
		\And
		\footnotesize CSIRO's Data61 (Australia)
		\AND
		\footnotesize Aalto University (Finland)
		\And
		\footnotesize EURECOM (France)
	}
]

\begin{abstract}

    Variational inference techniques based on inducing variables provide an elegant framework for scalable posterior estimation in \gls{GP} models. Besides enabling scalability, one of their main advantages over sparse approximations using direct marginal likelihood maximization is that they provide a robust alternative for point estimation of the inducing inputs, i.e.~the location of the inducing variables. In this work we challenge the common wisdom that optimizing the inducing inputs in the variational framework yields optimal performance. We show that, by revisiting old model approximations such as the fully-independent training conditionals endowed with powerful sampling-based inference methods, treating both inducing locations and \gls{GP} hyper-parameters in a Bayesian way can improve performance significantly. Based on stochastic gradient Hamiltonian Monte Carlo, we develop a fully Bayesian approach to scalable \gls{GP} and deep \gls{GP} models, and demonstrate its state-of-the-art performance through an extensive experimental campaign across several regression and classification problems.
\end{abstract}
\glsresetall
\section{INTRODUCTION}
\label{sec:intro}
Bayesian kernel machines based on \glspl{GP} combine the modeling flexibility of kernel methods with the ability to carry out sound quantification of uncertainty~\citep{Rasmussen06}.
Modeling and inference with \glspl{GP} have evolved considerably over the last few years with key contributions in the direction of scalability to virtually any number of datapoints and generality within automatic differentiation frameworks~\citep{GPflow17,Krauth17}.
This has been possible thanks to the combination of stochastic variational inference techniques with representations based on inducing variables \citep{Titsias09, gredilla2009inter,hensman13big}, random features~\citep{Rahimi08,Cutajar17,Gal16}, and structured approximations \citep{Wilson15,Wilson16}.
These advancements have now made \gps attractive  to a variety of applications and likelihoods~\citep{GPflow17,VanDerWilk17,bonilla-jmlr-2019}.

In this work, we focus on the variationally sparse \gp framework originally formulated by \citet{Titsias09} and later developed by \citet{hensman13big,Hensman15b} to scale up to large datasets via stochastic optimization.
In these formulations, the \gp prior is augmented with inducing variables (drawn from the same prior) and their posterior is approximated and estimated via variational inference. In contrast, the location of the inducing variables, which we refer to as the inducing inputs, are simply optimized along with covariance hyper-parameters.
In line with earlier evidence that Bayesian treatments of \gps are beneficial
~\citep{neal1997monte,barber1997gaussian,Murray10,FilipponeIEEETPAMI14}, posterior inference of the inducing variables \emph{jointly} with covariance hyper-parameters has been shown to improve performance \citep{Hensman15}.

Despite these significant insights with regards to the benefits of full Bayesian inference over latent variables in \gp models, the common practice is to optimize the inducing inputs, even in very recent \gp developments \citep{Havasi2018,shi2019sparse,giraldo2019fully}. 
In fact, the original work of \citet{Titsias09} advocates for a treatment of the inducing inputs as variational parameters to avoid overfitting.
Furthermore, later work concludes that point estimation of the inducing inputs through optimization of the variational objective is an `optimal' treatment \citep[][\S 3]{Hensman15}. As we will see in \cref{sec:previous-frameworks}, the  justification for inducing-input optimization in \citet{Hensman15}  relies on being able to optimize both the prior and the posterior, and therefore, contradicts the fundamental principles of Bayesian inference.
We summarize  previous works on inference methods for \gps in \cref{tab:methods}, which we will use for comparison in our experiments.
\begin{table*}[t]
    \begin{minipage}{.65\textwidth}
        \tikzexternaldisable
\centering
\captionof{table}{A summary of previous works on inference methods for \gps. $\mbtheta, \mbu, \mbZ$ refer to the \acrshort{GP} hyper-parameters, inducing variables and inducing inputs, respectively. (\xmark) indicates that variables are optimized.
    }
\label{tab:methods}
\scriptsize
\definecolor{color_bsgp}{rgb}{0.215686274509804,0.470588235294118,0.749019607843137}
\definecolor{color_fitc}{rgb}{0.658823529411765,0.709803921568627,0.0156862745098039}
\definecolor{color_svgp}{rgb}{1,0.650980,0.1686274}
\definecolor{color_mcsgp}{rgb}{0.52156862745098,0.403921568627451,0.596078431372549}
\definecolor{color_sghmc}{rgb}{1,0.27843137254902,0.298039215686275}
\definecolor{color_ipvi}{rgb}{0.360784313725490,0.662745098039216,0.0156862745098039}
\renewcommand{\tabcolsep}{0.5ex}
\begin{tabular}{p{.4cm}l ccc p{5cm}}
    \toprule
                                                                                                                                                                   &                      & \multicolumn{3}{c}{Inference} &                                                                                    \\
    \cmidrule{3-5}
                                                                                                                                                                   & Model                & $\mbtheta$                    & $\mbu$                  & $\mbZ$        & Reference                                \\
    \midrule
                                                                                                                                                                   & \textsc{mcmc-gp}     & \textsc{mcmc}                 & -                       & -             & \citet{neal1997monte,barber1997gaussian} \\
    ({\protect\tikz[baseline=-.65ex]\protect\draw[thick, color=color_svgp, fill=color_svgp, mark=*, mark size=2pt, line width=1.25pt] plot[] (-.0, 0)--(-0,0);})   & \textsc{svgp} ~      & \xmark                        & \textsc{vb}             & \xmark        & \citet{Hensman15b}                       \\
    ({\protect\tikz[baseline=-.65ex]\protect\draw[thick, color=color_fitc, fill=color_fitc, mark=*, mark size=2pt, line width=1.25pt] plot[] (-.0, 0)--(-0,0);})   & \textsc{fitc-svgp} ~ & \xmark                        & \textsc{vb} (heterosk.) & \xmark        & \citet{Titsias09}                        \\
    ({\protect\tikz[baseline=-.65ex]\protect\draw[thick, color=color_sghmc, fill=color_sghmc, mark=*, mark size=2pt, line width=1.25pt] plot[] (-.0, 0)--(-0,0);}) & \textsc{sghmc-dgp}   & \xmark                        & \textsc{mcmc}           & \xmark        & \citet{Havasi2018}                       \\
    ({\protect\tikz[baseline=-.65ex]\protect\draw[thick, color=color_ipvi, fill=color_ipvi, mark=*, mark size=2pt, line width=1.25pt] plot[] (-.0, 0)--(-0,0);})   & \textsc{ipvi-dgp}    & \xmark                        & \textsc{ip}             & \xmark        & \citet{Haibin2019}                       \\
    ({\protect\tikz[baseline=-.65ex]\protect\draw[thick, color=color_mcsgp, fill=color_mcsgp, mark=*, mark size=2pt, line width=1.25pt] plot[] (-.0, 0)--(-0,0);}) & \textsc{mcmc-svgp}   & \textsc{mcmc}                 & \textsc{mcmc}           & \xmark        & \citet{Hensman15}                        \\
    \midrule
    ({\protect\tikz[baseline=-.65ex]\protect\draw[thick, color=color_bsgp, fill=color_bsgp, mark=*, mark size=2pt, line width=1.25pt] plot[] (-.0, 0)--(-0,0);})   & \textsc{bsgp}        & \textsc{mcmc}                 & \textsc{mcmc}           & \textsc{mcmc} & This work                                \\
    \bottomrule
\end{tabular}
\tikzexternalenable
    \end{minipage}
    \hfill
    \setlength\figureheight{.16\textwidth}\setlength\figurewidth{.22\textwidth}\begin{minipage}{.33\textwidth}
        \tiny
        \pgfplotsset{every axis title/.append style={yshift=-2ex}}
        \includegraphics{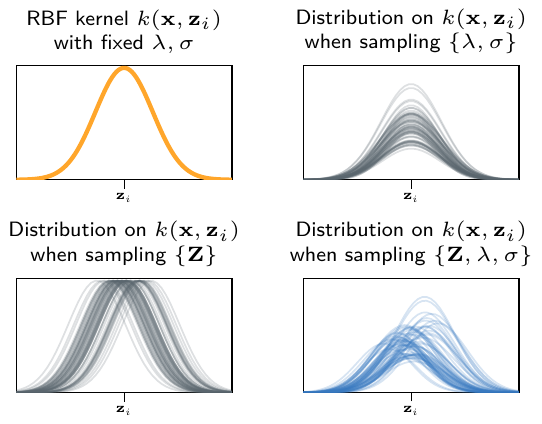}
        \captionof{figure}{Representation of the induced posterior distribution on the covariance function at location $\mbx$.}
        \label{fig:comparison_kernels}
    \end{minipage}
\end{table*}

Thus,  we revisit the role of the inducing inputs in \gp models and their treatment as variational parameters or even hyper-parameters.
Given their potential high dimensionality and that the typical number of inducing variables goes beyond hundreds/thousands \citep{shi2019sparse}, we argue that they should be treated simply as model variables and, therefore,  having priors and carrying out efficient posterior inference over them is an important---although challenging---problem.
An illustration of the richer modeling capabilities offered by treating inducing inputs in a Bayesian fashion is given in \cref{fig:comparison_kernels}.

\parhead{Contributions.}
Firstly,  we challenge the common wisdom that optimizing the inducing inputs in the variational framework yields optimal performance. We show that, by revisiting old model approximations such as the \acrlongpl{FITC} \citep[\acrshort{FITC}; see][]{quinonero2005unifying} endowed with powerful sampling-based inference methods, treating both inducing locations and \gp hyper-parameters in a Bayesian way can improve performance significantly. We describe the conceptual justification and the mathematical details of our general formulation  in \cref{sec:method} and \cref{sec:bsgp_in_practice}. We then demonstrate that our approach yields state-of-the-art performance across a wide range of competitive benchmark methods, large-scale datasets and a variety of \gp and deep \gp models (\cref{sec:experiments}).

\section{BAYESIAN SPARSE GAUSSIAN PROCESSES}\label{sec:method}

We are interested in supervised learning problems with $N$ input-label training pairs $\{\mbX, \mby\} \defeq \{(\mbx_i, y_i)\}_{i=1}^N$ , where we consider a conditional likelihood $p(\mby \g \mbf)$ and $\mbf$ is drawn from a zero-mean \gls{GP} prior with covariance function $k(\mbx, \mbx^\prime; \mbtheta)$ with hyper-parameters $\mbtheta$. Thus, we have that $p(\mbf) = \cN(\mathbold{0}, \mbK_{\mathrm{xx}|\mbtheta})$, where $\mbK_{\mathrm{xx}|\mbtheta}$ is the $N \times N$ covariance matrix obtained by evaluating $k(\mbx_i, \mbx_j; \mbtheta)$ over all input pairs $\{\mbx_i, \mbx_j\}$.
Inference in these types of models generally involves the costly $\bigO(N^3)$ operations to compute the inverse and log-determinant of the covariance matrix $\mbK_{\mathrm{xx}|\mbtheta}$.

\parhead{Full joint distribution of sparse approximations.}
Sparse \glspl{GP} are a family of approximate models that address the scalability issue by introducing a set of $M$ inducing variables $\mbu = (u_1, \ldots, u_M)$ at corresponding inducing inputs $\mbZ = \{\mbz_1, \ldots, \mbz_M\}$ such that $u_i = f(\mbz_i)$~\citep[see, e.g.,][]{quinonero2005unifying}.
These inducing variables are assumed to be drawn from the same \gls{GP} as the original process, yielding the joint prior $p(\mbf,\mbu) = p(\mbu) p(\mbf\g\mbu)$.
In the spirit of Bayesian modeling, any uncertainty in the covariance should be accounted for. 
Thinking of \gls{GP} hyper-parameters and inducing inputs as parameters of the covariance function, a distribution over these induces a distribution over the covariance function, which enriches the modeling capabilities of these models (see, e.g., \citet{jang2017scalable}).
We consider a general formulation where we
place priors $p_{\mbpsi}(\mbtheta)$ over covariance hyper-parameters and $p_{\mbxi}(\mbZ)$ over inducing inputs with hyper-parameters $\mbpsi,\mbxi$,
\begin{align}
    \label{eq:full-joint}
    p(\mbtheta, & \mbZ, \mbu, \mbf, \mby \g \mbX) =                                                                                                \\
                & = p_{\mbpsi}(\mbtheta)  p_\mbxi(\mbZ) p(\mbu \g \mbZ, \mbtheta)  p(\mbf \g \mbu, \mbX, \mbZ, \mbtheta) p(\mby \g \mbf)\nonumber,
    \end{align}
where $p(\mbu \g \mbZ, \mbtheta) =  \cN(\mathbf{0}, \mbK_{\mathrm{zz}|\mbtheta})$,
$ p(\mbf \g \mbu, \mbX, \mbZ, \mbtheta) = \cN(\mbK_{\mathrm{xz}|\mbtheta }\mbK_{\mathrm{zz}|\mbtheta}^{-1}\mbu, \mbK_{\mathrm{xx}|\mbtheta} -\mbK_{\mathrm{xz}|\mbtheta}\mbK_{\mathrm{zz}|\mbtheta}^{-1}\mbK_{\mathrm{xz}|\mbtheta}^\top)$.
The matrices $\mbK_{\mathrm{zz}|\mbtheta}, \mbK_{\mathrm{xz|\mbtheta}}$ denote the covariance matrices computed between points in $\mbZ$ and $\{\mbX, \mbZ\}$, respectively.
We assume a factorized likelihood $p(\mby \g \mbf) = \prod_{n=1}^N p(y_n \g f_n)$ and
make no assumptions about the other distributions. In this general formulation, approaches that do not consider priors over covariance hyper-parameters or inducing inputs correspond to improper uniform priors in \cref{eq:full-joint}.

\subsection{Scalable inference frameworks for GPs}
\label{sec:previous-frameworks}

Let $\mbPsi \defeq \{\mbu, \mbZ, \mbtheta\}$ be the variables whose posterior we wish to infer. Our main object of interest is the log joint marginal obtained by integrating out the latent variables $\mbf$ in \cref{eq:full-joint}, i.e., $\log p(\mby, \mbPsi \g \mbX) = \log \int_{\mbf}  p(\mby \g \mbf ) p(\mbf \g \mbPsi, \mbX)\text{d}\mbf  + \log p(\mbPsi)$. In particular,  we are interested in discussing approximations to this that decompose over observations, allowing the use of stochastic optimization techniques to scale up to large datasets.
In the literature of sparse \glspl{GP} \citep[see, e.g.,][]{Bauer2016,bui2017unifying}, two of the most influential methods for carrying out inference on such models are based on the \gls{VFE} framework \citep{Titsias09} and the \gls{FITC} framework \citep{Snelson2006}.

\parhead{VFE approximations.}
The key innovation in \citet{Titsias09} is the definition of the approximate posterior  $q(\mbf,\mbu) \defeq q(\mbu) p(\mbf\g \mbPsi, \mbX)$, where $q(\mbu)$ is the variational posterior, which yields the \gls{ELBO}
\begin{align} \label{eq:elbo}
     & p(\mathbf{y}\g\mbX, \mbZ, \mbtheta) \geq                                                                       \\
     & - \KL{q(\mbu)}{p(\mbu \g \mbZ, \mbtheta)} + \E_{q(\mbf,\mbu)} \log p(\mby \g \mbf) \defeq \Lelbo \nonumber \,.
\end{align}
We note that this approach does not incorporate priors over inducing inputs or hyper-parameters. Inference involves constraining $q(\mbu)$ to a parametric form and finding its parameters to optimize the \gls{ELBO}.
\citet{Titsias09} correctly argues that in the regression setting the variational approach to inducing variable approximations should be more robust to overfitting than a direct marginal likelihood maximization approach of traditional approximate models such as those described in \citet{quinonero2005unifying}. Indeed, if inducing inputs $\mbZ$ are optimized then the resulting \gls{ELBO} provides an additional regularization term \citep[see][\S 3 for details]{Titsias09}. However, as we shall see later, the benefits of being Bayesian about the inducing inputs and estimating their posterior distribution can be superior to those obtained by this regularization.

Restricting the form of $q(\mbu)$ is suboptimal, and \cite{Hensman15} proposes to sample from the optimal posterior approximation instead.
By applying Jensen's inequality  to bound the log joint marginal we obtain the following formulation,
\begin{align}
    \label{eq:logq}
     & \log p(\mby, \mbPsi \g \mbX) \geq                                                                                                               \\
     & \E_{p(\mbf \g \mbPsi, \mbX )}  \log p(\mby \g \mbf ) + \log p(\mbPsi) \defeq \log \widetilde{p}_{\textsc{vfe}}(\mby, \mbPsi \g \mbX) \nonumber.
\end{align}
This is the same expression derived in \cite{Hensman15}, although following  a different derivation showing that $\widetilde{p}_{\textsc{vfe}}$ indeed yields the optimal distribution under the \gls{VFE} framework of \cref{eq:elbo}.
However, \cite{Hensman15} argues that a Bayesian treatment of inducing inputs is unnecessary and concludes that the optimal prior is  $p(\mbZ) =  q(\mbZ) = \delta(\mbZ - \hat{\mbZ})$, where $\delta(\cdot)$ is Dirac's delta function and $\hat{\mbZ}$ is the set of inducing inputs that maximizes the \gls{ELBO} \citep[][\S 3]{Hensman15}. We find such a justification flawed as it contradicts the fundamental principles of Bayesian inference. Indeed, the derivation by \citet{Hensman15} relies on minimizing both sides of the KL term in \cref{eq:elbo}, allowing for a `free-form' optimization of the prior, which  ultimately negates the necessity  of all prior choices and defeats the purpose of a Bayesian treatment.

\parhead{FITC approximations.}
As an alternative, we can approximate the log joint of \cref{eq:full-joint} by imposing independence in the conditional distribution  \citep[see][for details]{quinonero2005unifying}, i.e.,~  parameterizing $p(\mbf\g\mbPsi,\mbX) = \cN\left(\mbK_{\mathrm{xz}|\mbtheta}\mbK_{\mathrm{zz}|\mbtheta}^{-1}\mbu, \diag\left[\mbK_{\mathrm{xx}|\mbtheta} -\mbK_{\mathrm{xz}|\mbtheta}\mbK_{\mathrm{zz}|\mbtheta}^{-1}\mbK_{\mathrm{xz}|\mbtheta}^\top\right]\right)$,
\begin{align}
     & \log p(\mby, \mbPsi \g \mbX)                                                                                                                                                        \approx \nonumber                                                                                                                                                                      \\
     & \underbrace{\sum\nolimits_{n=1}^N \log \E_{p(f_n \g \mbPsi, \mbX )} \left[ p(y_n \g f_n )\right] + \log p(\mbPsi)}_{ \defeq \log \tilde{p}_{\textsc{fitc}}(\mby, \mbPsi \g \mbX)} .
    \label{eq:logptilde}
\end{align}
This same formulation of the \gls{FITC} objective can be also been obtain by modifying the likelihood or the prior rather then the conditional distribution \citep{Snelson2006, Titsias09, Bauer2016}.

We now see that, when considering i.i.d.~conditional likelihoods, both approximations, $\log \tilde{p}_{\textsc{VFE}}$ and $\log \tilde{p}_{\textsc{FITC}}$, yield objectives that decompose on the observations, enabling scalable inference methods. In particular, we aim to sample from the posterior over all the latent variables using scalable approaches such as \acrlong{SGHMC} \citep[\acrshort{SGHMC};][]{Cheni2014}. The main question is what approach should be preferred and how they relate to their optimization counterparts.

\subsection{Sampling with VFE or FITC?}\label{sec:discussion_obj}
We will show in \cref{sec:experiments} that our proposal that samples from the posterior according to \cref{eq:logptilde} consistently outperforms that in \cref{eq:logq}.
To understand why the \gls{FITC} objective makes sense we need to go back to the original work of \citet{Titsias09,titsias2009techr} and the seminal work of \citet{quinonero2005unifying}.
Indeed, \cite{titsias2009techr} shows that, in the standard regression case with homoskedastic observation noise, \gls{VFE} yields exactly the same predictive posterior as the \gls{PP} approximation \citep{seeger2003fast}, which is referred to as the \gls{DTC} approximation.  Despite this equivalence, as highlighted in \cite{Titsias09}, the main difference is that the \gls{VFE} framework provides a more robust approach to hyper-parameter estimation as the resulting \gls{ELBO} corresponds to a regularized marginal likelihood of the \gls{DTC} approach. Nevertheless, the \gls{DTC}/\gls{PP}, and consequently the \gls{VFE}, predictive distribution has been shown to be less accurate than the \gls{FITC} approximation \citep{Titsias09,quinonero2005unifying,snelson2007flexible}.
Effectively, as described by \citet{quinonero2005unifying}, the \gls{VFE}'s solution (which is the same as \gls{DTC}'s) can be understood as considering a deterministic conditional prior $p(\mbf \g \mbu)$, i.e.~a conditional prior with zero variance.

Consequently, the main reason for the superior performance of \gls{VFE} in earlier approaches, despite providing a less accurate predictive posterior than \gls{FITC}'s, was that inducing input estimation was less prone to overfitting due to the use of the variational objective, which provided an extra regularization term.
However, by placing priors over the inducing inputs $\mbZ$ as well as over covariance hyper-parameters (as we propose in this work),  regularization  over these parameters   becomes unnecessary.
For this reason, we expect  the \emph{log of the expectation} in \cref{eq:logptilde} to provide more accurate results than the \emph{expectation of the log} in \cref{eq:logq}.
Finally, it is important to point out that a variational formulation equivalent to \gls{FITC} has also been proposed \citep[see][App.~C]{titsias2009techr}. Our experimental evaluation, also assesses the benefits of our method with respect to that approach.
Full details of this analysis can be found in the supplement.

\begin{figure}[t]
    \centering
    \pgfplotsset{every axis title/.append style={yshift=-2ex}}
    \setlength\figureheight{.22\textwidth}\setlength\figurewidth{.20\textwidth}
    \centering\tiny
    \begin{subfigure}[t]{0.25\textwidth}
        \centering
        \includegraphics{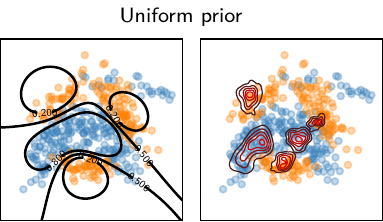}
    \end{subfigure}\begin{subfigure}[t]{0.25\textwidth}
        \centering
        \includegraphics{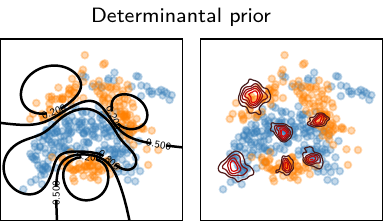}
    \end{subfigure}\\[1ex]
    \begin{subfigure}[b]{0.25\textwidth}
        \centering
        \includegraphics{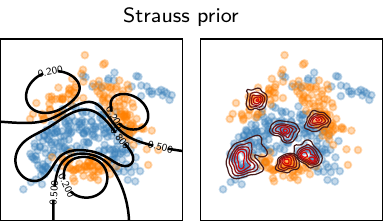}
    \end{subfigure}\begin{subfigure}[b]{0.25\textwidth}
        \centering
        \includegraphics{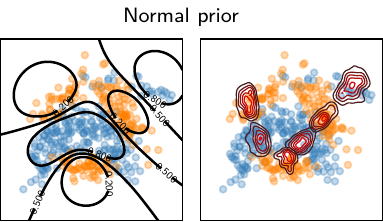}
    \end{subfigure}
    \caption{Illustration of a binary classification task on the \textsc{banana} dataset.
        \emph{Left}: the decision bounds of the average classifier.
        \emph{Right}: the posterior marginals of the inducing inputs. }
    \label{fig:2d_banana}
\end{figure}\section{PRACTICAL CONSIDERATIONS AND EXTENSION TO DEEP GPs}\label{sec:bsgp_in_practice}
In this section we describe practical considerations in our \gls{BSGP} framework, including inference techniques, prior choices and extensions to deep Gaussian processes.
Recalling that $\mbPsi=\{\mbtheta, \mbu, \mbZ\}$ represents the set of variables to infer and, using \cref{eq:logptilde}, their posterior  can be obtained as
\begin{align}
    \label{eq:logposterior_sghmc}
    \log p(\mbPsi\g\mby,\mbX) = & \log \mathbb{E}_{p(\mbf\g\mbPsi,\mbX)} p(\mby\g\mbf) +\log p(\mbu\g\mbtheta,\mbZ) + \nonumber \\
                                & \log p_\mbxi(\mbZ) + \log p_\mbpsi(\mbtheta) - \log C.
\end{align}
We use \gls{MCMC} techniques, in particular  \gls{SGHMC} \citep{Cheni2014,Havasi2018}, to obtain samples from the intractable $p(\mbPsi\g\mby,\mbX)$.
Unlike \gls{HMC}, which requires computing the exact gradient $\nabla \log p(\mbPsi\g\mby,\mbX)$ and the exact unnormalized posterior to evaluate the acceptance \citep{neal10},
\gls{SGHMC} obtains samples from the posterior with stochastic gradients and without evaluating the Metropolis ratio (see supplement for details).
With a factorized likelihood $p(\mby|\mbf)$ and an energy function  $U(\mbPsi) = -\log p(\mbPsi\g\mby,\mbX) + \log C$, we sample  \cref{eq:logposterior_sghmc} over minibatches of data.

\subsection{Choosing priors}\label{sec:prior_choice}
\label{sec:priors}

Next, we discuss prior choices for the inducing inputs and covariance hyper-parameters. The inducing inputs $\mbZ$ support the sparse Gaussian process interpolation, which motivates matching the inducing prior to the data distribution $p(\mbX)$. We begin by proposing a simple Normal (N) prior
$
    p_N(\mbZ)  = \prod_{j=1}^M \cN( \mbz_j | \mathbf{0}, \mbI),
$
which matches the mean and variance of the normalized data distribution, and favors inducing inputs toward the baricenter of  the data inputs.

We also explore two priors based on point processes, which consider distributions over point sets \citep{gonzalez2016spatio}. Point processes can induce repulsive effects penalizing configurations where inducing points are clumped together. The \gls{DPP}, defined through
$
    p_{D}(\mbZ)  \propto \det \mbK_{\mathrm{zz}|\mbtheta}\,, $
relates the probability of inducing inputs to the volume of space spanned by the covariance \citep{lavancier2014}. \gls{DPP} is a repulsive point process, which gives higher probabilities to input diversity, controlled by the hyper-parameters $\mbxi \equiv \mbtheta$.
We then consider the Strauss process \citep[see e.g.][]{vere-jones-book-2003, Strauss1975},
$
    p_S(\mbZ)  \propto \lambda^M \gamma^{\sum_{\mbz,\mbz' \in \mbZ} \delta(|\mbz - \mbz'| < r) },
$
where $\lambda > 0$ is the intensity, and $0 < \gamma \le 1$ is the repulsion coefficient which decays the prior as a function of the number of input pairs that are within distance $r$. The Strauss prior (S) tends to maintain the minimum distance between inducing inputs, parameterized by $\mbxi = (\lambda,\gamma,r)$.
We finally consider an uninformative uniform  prior~(U), $\log p_U(\mbZ) = 0$, which effectively provides no contribution to the evaluation of the posterior.
To gain insights on the choice of these priors, we set up a comparative analysis on the \name{banana} dataset (\cref{fig:2d_banana}).
We observe that the posterior densities on the inducing inputs are multimodal and highly non-Gaussian, further confirming the necessity of free-form inference.
Both Strauss and \gls{DPP}-based priors encourage configurations where the inducing inputs are evenly spread.
The Normal and Uniform priors, instead, focus exclusively on aligning the inducing inputs in a way that is sensible to accurately model the intricate classification boundary between the classes.
This insight is confirmed by our the extensive experimental validation in \cref{sec:experiments}.

\paragraph{Prior on covariance hyper-parameters.}
Choosing priors on the hyper-parameters has been discussed in previous works on Bayesian inference for \glspl{GP} \citep[see e.g.][]{FilipponeIEEETPAMI14}.
Throughout this paper, we use the RBF covariance with \gls{ARD}, marginal variance $\sigma$ and independent lengthscales $\lambda_i$ per feature \citep{Mackay94}.
On these two hyper-parameters we place a lognormal prior with unit variance and means equal to 1 and 0.05 for $\mblambda$ and $\sigma$, respectively.

\subsection{Extension to deep Gaussian processes}
\gls{BSGP} can be easily extended to \gls{DGP} models \citep{Damianou13}, where we compose $L$ sparse \gls{GP} layers.
Each layer is associated with a set of inducing inputs $\mbZ_\ell$, inducing variables $\mbu_\ell$ and hyper-parameters $\mbtheta_\ell$ \citep{Salimbeni17}.
In our notation $\mbPsi = \{\mbPsi_\ell\}_{\ell=1}^L = \{\mbZ_\ell, \mbu_\ell, \mbtheta_\ell\}_{\ell=\ell}^L$.
The joint distribution is
\begin{equation}
    p\left(\mby, \mbPsi \right) =  p\left(\mby\g\mbf_L\right) \prod\nolimits_{\ell=1}^{L} p\left(\mbf_\ell\g \mbf_{\ell-1}, \mbPsi_\ell\right)p(\mbPsi_\ell),
\end{equation}
where we omit dependency on $\mbX$.
In contrast to the `shallow' joint distribution in \cref{eq:full-joint} and posterior in \cref{eq:logposterior_sghmc}, the `hidden' layers $\mbf_\ell$ are marginalized with sampling and propagated up to the final layer $L$ \citep{Salimbeni17}, which can be marginalized exactly if the likelihood is Gaussian or by quadrature \citep{Hensman15b}. Full details and derivations for this more general case can be found in the supplement.

\section{EXPERIMENTS}\label{sec:experiments}
\begin{figure*}[t]
    \pgfplotsset{every axis title/.append style={yshift=-2ex}}
    \centering\tiny
    \setlength\figureheight{.15\textwidth}
    \setlength\figurewidth{.2\textwidth}
    \begin{subfigure}[t]{0.56\textwidth}
        \centering
        \includegraphics{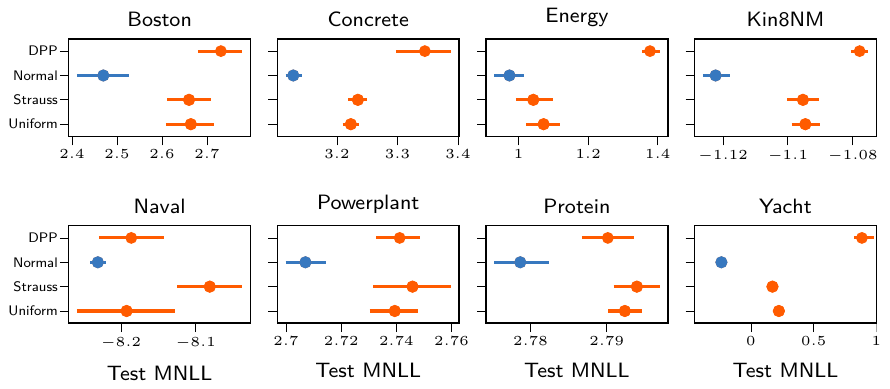}\\[2ex]
        \tikzexternaldisable
\definecolor{color0}{rgb}{1,0.356862745098039,0}
\definecolor{color1}{rgb}{0.215686274509804,0.470588235294118,0.749019607843137}
\begin{tabular}{p{.3cm}l}
    \toprule
    {\protect\tikz[baseline=-1ex]\protect\draw[thick, color=color0, fill=color0, mark=*, mark size=1.5pt, line width=1.25pt] plot[] (-.0, 0)--(.25,0)--(-.25,0);} & \textsf{\hsc{bsgp} with Determinantal Point Process prior (\hsc{dpp}), Strauss process prior, Uniform prior} \\
    {\protect\tikz[baseline=-1ex]\protect\draw[thick, color=color1, fill=color1, mark=*, mark size=1.5pt, line width=1.25pt] plot[] (-.0, 0)--(.25,0)--(-.25,0);} & \textsf{\hsc{bsgp} with Normal prior} \\
    \bottomrule
\end{tabular}
\tikzexternalenable
    \end{subfigure}
    \setlength\figurewidth{.22\textwidth}\unskip\ \vrule\
    \begin{subfigure}[t]{0.4\textwidth}
        \centering
        \includegraphics{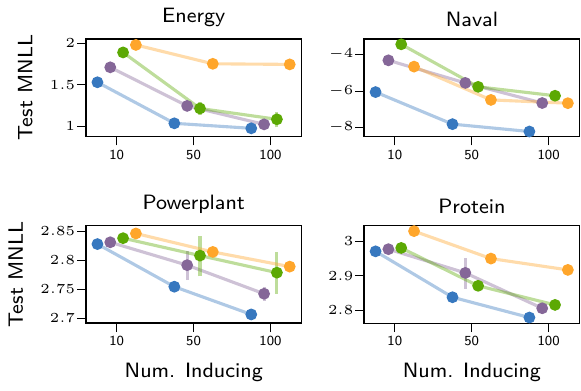}\\[1ex]
        \tikzexternaldisable
\definecolor{color3}{rgb}{1,0.650980392156863,0.168627450980392}
\definecolor{color2}{rgb}{0.36078431372549,0.662745098039216,0.0156862745098039}
\definecolor{color0}{rgb}{0.215686274509804,0.470588235294118,0.749019607843137}
\definecolor{color1}{rgb}{0.52156862745098,0.403921568627451,0.596078431372549}
\begin{tabular}{p{.3cm}l}
    \toprule
    {\protect\tikz[baseline=-1ex]\protect\draw[thick, color=color3, fill=color3, mark=*, mark size=1.5pt, line width=1.25pt] plot[] (-.0, 0)--(.25,0)--(-.25,0);} & \textsf{Gaussian $q(\mbu)$ [\hsc{svgp}]}                            \\
    {\protect\tikz[baseline=-1ex]\protect\draw[thick, color=color2, fill=color2, mark=*, mark size=1.5pt, line width=1.25pt] plot[] (-.0, 0)--(.25,0)--(-.25,0);} & \textsf{Free form $p(\mbu\g\mby)$ [\hsc{sghmc-dgp}]}                \\
    {\protect\tikz[baseline=-1ex]\protect\draw[thick, color=color1, fill=color1, mark=*, mark size=1.5pt, line width=1.25pt] plot[] (-.0, 0)--(.25,0)--(-.25,0);} & \textsf{Free form $p(\mbu,\mbtheta\g\mby)$}                            \\
    {\protect\tikz[baseline=-1ex]\protect\draw[thick, color=color0, fill=color0, mark=*, mark size=1.5pt, line width=1.25pt] plot[] (-.0, 0)--(.25,0)--(-.25,0);} & \textsf{Free form $p(\mbu,\mbtheta, \mbZ\g\mby)$ [\hsc{\textbf{bsgp - this work}}]} \\
    \bottomrule
\end{tabular}
\tikzexternalenable
    \end{subfigure}
    \caption{ {\textit{Left:}} analysis of different priors on inducing locations for \gls{BSGP} on the UCI benchmarks:
        \glsentryfull{DPP},
        Strauss process,
        uniform. {\textit{Right:}} ablation study on the effect of performing posterior inference on different sets of variables.
        From \name{svgp}, where the posterior is constrained to be Gaussian and the remaining parameters are point-estimated, to our proposal \gls{BSGP}, where we infer a free-form posterior for all $ \mbPsi = \{\mbu, \mbtheta, \mbZ\}$.
        We refer the reader to  \cref{tab:methods} for  details on the methods (colors are matched).
    }
    \label{fig:ablation_study}
\end{figure*}
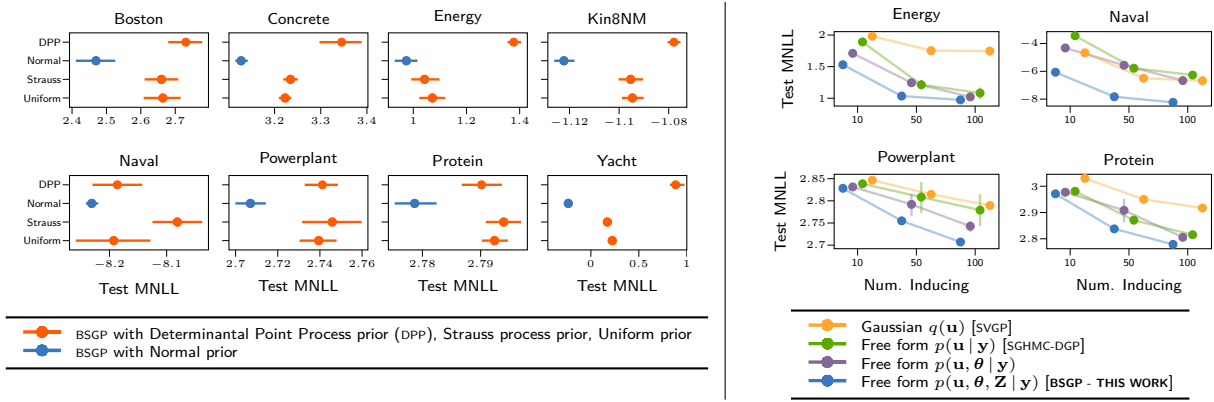

In this section, we provide empirical evidence that our \gls{BSGP} outperforms previous inference/optimization approaches on shallow and deep \gls{GP}s.
We use eight of the classic UCI benchmark datasets with standardized features and split
into eight folds with 0.8/0.2 train/test ratio.
We train the competing models for 10,000 iterations with \name{adam} \citep{Kingma2015b}, step size of 0.01 and a minibatch of 1,000 samples.
The sampling methods are evaluated based on 256 samples collected after optimization. Following previous works \citep[e.g.][]{Rasmussen06,Havasi2018,Haibin2019}, in order to evaluate and compare the full predictive posteriors we compute the \gls{MNLL} on the test set (\textsc{rmse}s are reported in the supplement for reference).

\subsection{Prior analysis and ablation study}
We start our empirical analysis  with a comparative evaluation of the priors on inducing inputs
described in~\cref{sec:prior_choice}:
\gls{DPP}, Normal, Strauss and Uniform.
We run our inference procedure on a shallow \gls{GP} with 100 inducing points and we report the results in \cref{fig:ablation_study} (left).
The results show that the Normal prior consistently outperforms the others.
The uniform and Strauss priors behave similarly, while the \name{dpp} prior is consistently among the worst.
We argue that the repulsive nature of the point process priors (\name{dpp}, particularly), although grounded on the intuition of covering the input space more evenly, constrains the smoothness of the functions up to the point that they become too simple to accurately model the data.
With this, we select the Gaussian prior for the remaining experiments.

We now study the benefits of a Bayesian treatment of the inducing variables, inducing inputs, and hyper-parameters with an ablation study.
Using the same setup as before, we start with the baseline of \name{svgp} \citep{hensman13big,Hensman15b}, where the posterior on $\mbu$ is approximated using a Gaussian and $\mbZ, \mbtheta$ are optimized.
We then incrementally add parameters to the list of variables that are sampled rather than optimized: only $\mbu$ (equivalent to \name{sghmc-dgp}, \citep{Havasi2018}), then $\{\mbu, \mbtheta\}$ and finally, our proposal, $\{\mbu, \mbtheta, \mbZ\}$.
This experiment is repeated for different number of inducing points (10, 50 and 100).
\cref{fig:ablation_study} (right) reports a summary of these results (full comparison in the supplement).
This plot shows that each time we carry out free-form posterior inference on a bigger set of parameters rather than optimization, performance is enhanced, and our proposal outperforms previous approaches. 
\setlength\figureheight{.24\textwidth}
\setlength\figurewidth{.125\textwidth}
\begin{figure*}[t!]
    \begin{minipage}[bt]{.65\textwidth}
        \pgfplotsset{every axis title/.append style={yshift=-2ex}}
        \centering\tiny
        \includegraphics{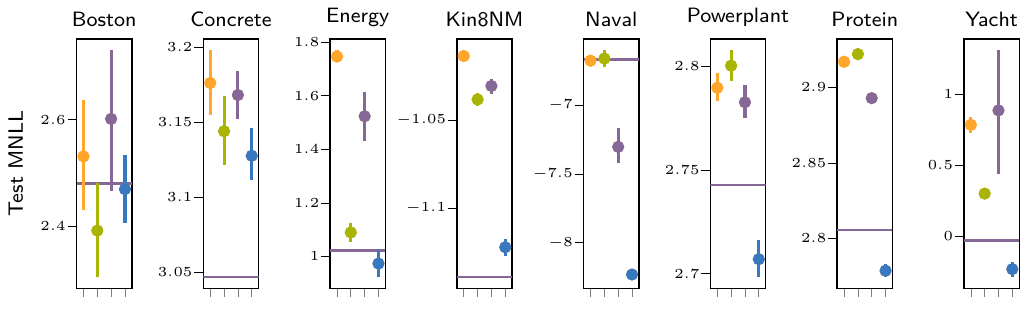}
        \tikzexternaldisable
\definecolor{color3}{rgb}{0.215686274509804,0.470588235294118,0.749019607843137}
\definecolor{color2}{rgb}{0.658823529411765,0.709803921568627,0.0156862745098039}
\definecolor{color0}{rgb}{1,0.650980392156863,0.168627450980392}
\definecolor{color1}{rgb}{0.52156862745098,0.403921568627451,0.596078431372549}
\begin{tabular}{cl}
    \toprule
    {\protect\tikz[baseline=-1ex]\protect\draw[thick, color=color0, fill=color0, mark=*, mark size=1.5pt, line width=1.25pt] plot[] (-.0, 0)--(.25,0)--(-.25,0);} & \textsf{Inference of $\mbu$ on the variational objective \citep[\hsc{svgp}~--][]{Hensman15b} }\\
    {\protect\tikz[baseline=-1ex]\protect\draw[thick, color=color2, fill=color2, mark=*, mark size=1.5pt, line width=1.25pt] plot[] (-.0, 0)--(.25,0)--(-.25,0);} & \textsf{Inference of $\mbu$ with heteroskedastic likelihood (equivalent to \hsc{fitc}) \citep[\hsc{fitc-svgp}~--][]{Titsias09}} \\
    {\protect\tikz[baseline=-1ex]\protect\draw[thick, color=color1, fill=color1, mark=*, mark size=1.5pt, line width=1.25pt] plot[] (-.0, 0)--(.25,0)--(-.25,0);} & \textsf{MCMC inference of $\mbu, \mbtheta$ on the variational objective \citep[\hsc{mcmc-svgp}~--][]{Hensman15}} \\
    {\protect\tikz[baseline=-1ex]\protect\draw[thick, color=color1, fill=color1, mark size=1.5pt, line width=1.25pt] plot[] (-.0, 0)--(.25,0)--(-.25,0);} & \textsf{MCMC inference of $\mbu, \mbtheta$ on the marginal likelihood \textit{(log of expectation)}} \\
    {\protect\tikz[baseline=-1ex]\protect\draw[thick, color=color3, fill=color3, mark=*, mark size=1.5pt, line width=1.25pt] plot[] (-.0, 0)--(.25,0)--(-.25,0);} & \textsf{MCMC inference of $\mbu, \mbtheta, \mbZ$ on the marginal likelihood [\hsc{\textbf{bsgp - this work}}]} \\
    \bottomrule
\end{tabular}
\tikzexternalenable
        \captionof{figure}{Analysis of different choices of objectives when used for optimization and sampling.
            We refer the reader to \cref{tab:methods} for a description of the methods.
        }
        \label{fig:ablation_study_objectives}
    \end{minipage}\hfill\setlength\figureheight{.22\textwidth}\setlength\figurewidth{.35\textwidth}\begin{minipage}[bt]{.32\textwidth}
        \centering\tiny
        \includegraphics{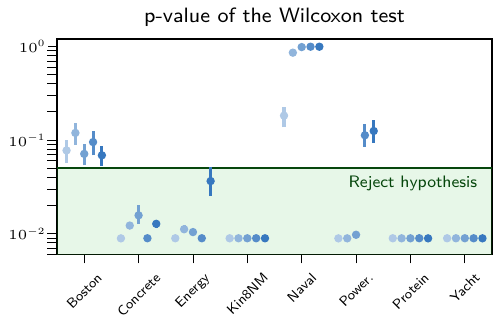}
        \tikzexternaldisable
        \definecolor{color4}{rgb}{0.215686274509804,0.470588235294118,0.749019607843137}
        \definecolor{color5}{rgb}{0.686800573888092,0.789383070301291,0.901434720229555}
        \captionof{figure}{p-values of the hypothesis test that \gls{BSGP} with \gls{VFE} objective is better than \gls{BSGP} with \gls{FITC} objective; depth of the \gls{DGP} from 1
            ({\protect\tikz[baseline=-.65ex]\protect\draw[thick, color=color5, fill=color5, mark=*, mark size=2pt, line width=1.25pt] plot[] (-.0, 0)--(-0,0);})
            to 5
            ({\protect\tikz[baseline=-.65ex]\protect\draw[thick, color=color4, fill=color4, mark=*, mark size=2pt, line width=1.25pt] plot[] (-.0, 0)--(-0,0);}).
            For models with $\text{p-values} < 0.05$, we reject the hypothesis.}
        \label{fig:wilcoxon_test}
        \tikzexternalenable
    \end{minipage}
\end{figure*}

\setlength\figureheight{.16\textwidth}
\setlength\figurewidth{.17\textwidth}
\begin{figure}[b]
    \centering\tiny
    \includegraphics{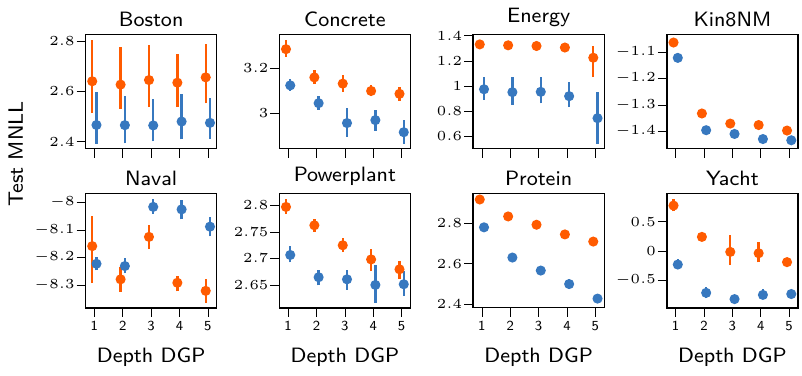}
    \\[-2.5ex]
    \tikzexternaldisable
\definecolor{color0}{rgb}{1,0.356862745098039,0}
\definecolor{color1}{rgb}{0.215686274509804,0.470588235294118,0.749019607843137}
\captionof{figure}{\gls{BSGP} with the two different objectives for the sampler: \gls{FITC} ({\protect\tikz[baseline=-.75ex]\protect\draw[thick, color=color1, fill=color1, mark=*, mark size=1.5pt, line width=1.25pt] plot[] (-.0, 0)--(.25,0)--(-.25,0);}) and \gls{VFE} ({\protect\tikz[baseline=-.75ex]\protect\draw[thick, color=color0, fill=color0, mark=*, mark size=1.5pt, line width=1.25pt] plot[] (-.0, 0)--(.25,0)--(-.25,0);}).
}
\tikzexternalenable\label{fig:bsgp-comparison_obj}
\end{figure}

\subsection{Choosing the objective: VFE vs FITC}
In  \cref{sec:method}
we discussed the role of the marginal and the \glsentryfull{VFE} objective when used for optimization and for sampling.
In \cref{fig:ablation_study_objectives} we support the discussion with  empirical results.
The baseline is \name{svgp}, for which the inference is approximate (Gaussian) and performed on the variational objective.
\citet[App.~C]{titsias2009techr} also considers  a \gls{VFE} formulation of \gls{FITC} which corresponds to a \gls{GP} regression with heteroskedastic noise variance.
The likelihood needs to be augmented to handle heteroskedasticity, but inference can be carried out exactly on the variational objective.
For these two methods, $\{\mbtheta, \mbZ\}$ are optimized.
We also test \name{mcmc-svgp}, the model proposed by \citet{Hensman15}, implemented in {GPflow} \citep{GPflow17} with the same suggested experimental setup.
This experiment indicates that having a free-form posterior on $\mbu,\mbtheta$ sampled from the variational objective does not dramatically improve on the exact Gaussian approximation of the \gls{FITC} model, with both of them delivering superior performance with respect to \name{svgp}.
In the same setup of \citet{Hensman15} ($\mbu,\mbtheta$ sampled and $\mbZ$ optimized), we look at the effect of swapping the {\it expectation of log} with the {\it log of expectation} (which effectively means moving from the \gls{VFE} objective to  \gls{FITC}); on the contrary, here we observe a significant increase in performance when using the latter, further confirming the discussion of the objectives in \cref{sec:discussion_obj}.
We finally conclude this section with an experiment where we try both objectives on our proposed \gls{BSGP} and also different depths of the \gls{DGP} (\cref{fig:bsgp-comparison_obj}).
Using the Wilcoxon signed-rank test \citep{Wilcoxon1945}, we test the null hypothesis of \gls{VFE} objective being better than the proposed \gls{FITC}.
\cref{fig:wilcoxon_test} shows that,
for the majority of the cases, this can be rejected ($p<0.05$).

\setlength\figureheight{.28\textwidth}
\setlength\figurewidth{.245\textwidth}
\pgfplotsset{every y tick label/.append style={font=\fontsize{5}{5}\selectfont}}
\begin{figure*}[t]
    \begin{subfigure}[t]{.775\textwidth}
        \centering\tiny
        \includegraphics{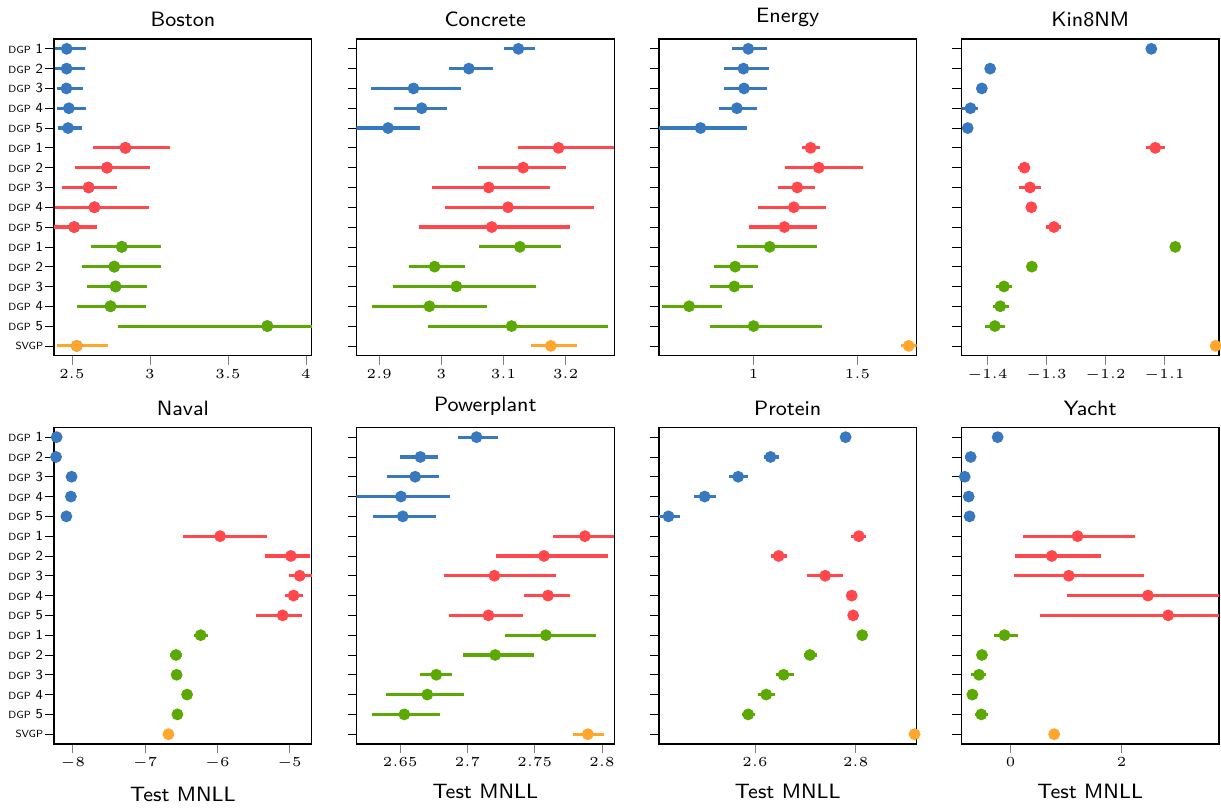}
        \label{fig:mnll-benchmark-comparison}
    \end{subfigure}\begin{subfigure}[b]{.19\textwidth}
        \centering\tiny
        \tikzexternaldisable
\definecolor{color0}{rgb}{1,0.650980392156863,0.168627450980392}
\definecolor{color1}{rgb}{1,0.278431,0.2980392}
\definecolor{color2}{rgb}{0.3607843,0.66274509,0.01568627}
\definecolor{color3}{rgb}{0.2156862,0.4705882,0.74901960}
\begin{tabular}{cl}
    \toprule
    {\protect\tikz[baseline=-1ex]\protect\draw[thick, color=color3, fill=color3, mark=*, mark size=1.5pt, line width=1.25pt] plot[] (-.0, 0)--(.25,0)--(-.25,0);} & \textsf{BSGP (\textbf{This work})} \\[1ex]
    {\protect\tikz[baseline=-1ex]\protect\draw[thick, color=color1, fill=color1, mark=*, mark size=1.5pt, line width=1.25pt] plot[] (-.0, 0)--(.25,0)--(-.25,0);} & \textsf{IPVI-DGP}                  \\
                                                                                                                                                                  & \textsf{\citep{Haibin2019}}        \\[1ex]
    {\protect\tikz[baseline=-1ex]\protect\draw[thick, color=color2, fill=color2, mark=*, mark size=1.5pt, line width=1.25pt] plot[] (-.0, 0)--(.25,0)--(-.25,0);} & \textsf{SGHMC-DGP}                 \\
                                                                                                                                                                  & \textsf{\citep{Havasi2018} }       \\[1ex]
    {\protect\tikz[baseline=-1ex]\protect\draw[thick, color=color0, fill=color0, mark=*, mark size=1.5pt, line width=1.25pt] plot[] (-.0, 0)--(.25,0)--(-.25,0);} & \textsf{SVGP}                      \\
                                                                                                                                                                  & \textsf{\citep{Hensman15b}}        \\
    \bottomrule
\end{tabular}
\tikzexternalenable\\[10ex]
        \includegraphics{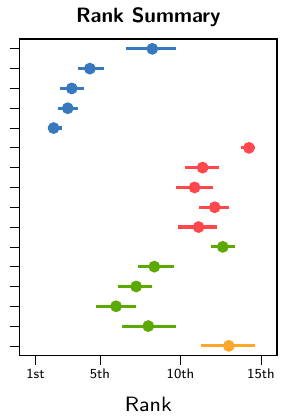} \vspace{-0ex}\par
        \label{fig:mnll-benchmark-rank}
    \end{subfigure}
    \caption{Test \gls{MNLL} on UCI regression benchmarks (the error bars represent the $95\%$CI). The lower \gls{MNLL} (i.e. to the left), the better. The number on the right of the method's name refers to the depth of the \gls{DGP}. \emph{Bottom right}: Rank summary of all methods. }
    \label{fig:uci8}
\end{figure*}

\setlength\figureheight{.215\textwidth}
\setlength\figurewidth{.29\textwidth}
\begin{figure}[b!]
    \centering\tiny
    \pgfplotsset{every axis title/.append style={yshift=-1ex}}
    \includegraphics{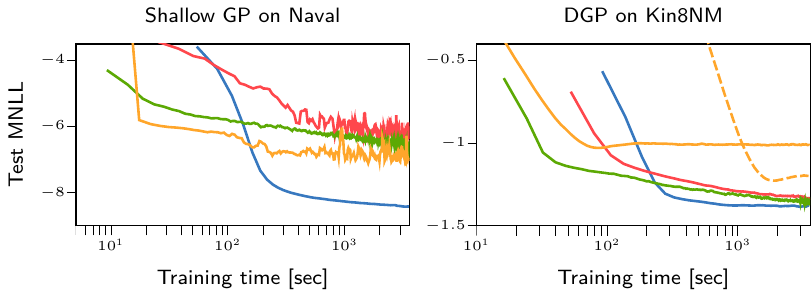}
    \tikzexternaldisable
\definecolor{color0}{rgb}{1,0.650980392156863,0.168627450980392}
\definecolor{color1}{rgb}{1,0.278431,0.2980392}
\definecolor{color2}{rgb}{0.3607843,0.66274509,0.01568627}
\definecolor{color3}{rgb}{0.2156862,0.4705882,0.74901960}
\setlength{\tabcolsep}{5pt}
\begin{tabular}{llll}
    \toprule
    {\protect\tikz[baseline=-1ex]\protect\draw[thick, color=color3, fill=color3, mark=*, mark size=1.5pt, line width=1.25pt] plot[] (-.0, 0)--(.25,0)--(-.25,0);}  \textsf{BSGP} &
    {\protect\tikz[baseline=-1ex]\protect\draw[thick, color=color1, fill=color1, mark=*, mark size=1.5pt, line width=1.25pt] plot[] (-.0, 0)--(.25,0)--(-.25,0);}  \textsf{IPVI-DGP}                  &
    {\protect\tikz[baseline=-1ex]\protect\draw[thick, color=color2, fill=color2, mark=*, mark size=1.5pt, line width=1.25pt] plot[] (-.0, 0)--(.25,0)--(-.25,0);}  \textsf{SGHMC-DGP}                 &
    {\protect\tikz[baseline=-1ex]\protect\draw[thick, color=color0, fill=color0, mark=*, mark size=1.5pt, line width=1.25pt] plot[] (-.0, 0)--(.25,0)--(-.25,0);}  \textsf{SVGP}                      \\
    \bottomrule
\end{tabular}
\tikzexternalenable
    \captionof{figure}{Comparison of test \gls{MNLL} as function of training time. The dashed line on the right hand side plot corresponds to \textsc{svgp} with $M=1000$ inducing points.}
    \label{fig:time-comparison}
    \end{figure}
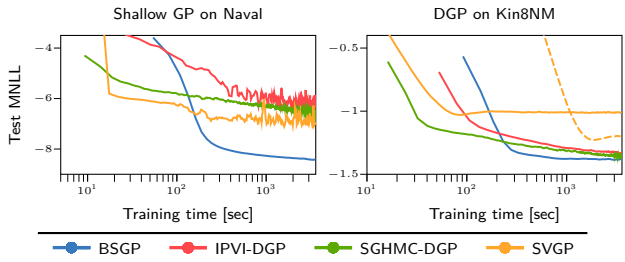

\subsection{Deep Gaussian processes on UCI benchmarks}
We now report results on \glspl{DGP}.
We compare against two current state-of-the-art deep \gls{GP} methods, \textsc{sghmc-dgp} \citep{Havasi2018} and \textsc{ipvi-dgp} \citep{Haibin2019}, and against the shallow \textsc{svgp} baseline \citep{Hensman15b}.
For a faithful comparison with \textsc{ipvi-dgp} we follow the recommended parameter configurations\footnote{We use the \textsc{ipvi-dgp} implementation available at \href{https://github.com/HeroKillerEver/ipvi-dgp}{\tt github.com/HeroKillerEver/ipvi-dgp}}.
Using a standard setup, all models share $M=100$ inducing points, the same RBF covariance with \gls{ARD} and, for \gls{DGP}, the same hidden dimensions (equal to the input dimension $D$).
\cref{fig:uci8} shows the predictive test \gls{MNLL} mean and $95\%$ CI over the different folds over the UCI datasets, and also includes rank summaries. The proposed method clearly outperforms competing deep and shallow \glspl{GP}.
The improvements are particularly evident on \textsc{naval}, a dataset known to be challenging to improve upon.
Furthermore, the deeper models perform consistently better or on par with the shallow version, without incurring in any measurable overfitting even on small or medium sized datasets (see \textsc{boston} and \textsc{yacht}, for example).

\parhead{Computational efficiency.}
Similarly to the baseline algorithms, each training iteration of \gls{BSGP} involves the computation of the inverse covariance with complexity $\bigO(M^3)$.
In \cref{fig:time-comparison} we compare the three main competitors with \gls{BSGP} trained for a fixed training time budget of one hour for a shallow \gls{GP} and a 2-layer \gls{DGP}.
The experiment is repeated four times on the same fold and the results are then averaged.
Each run is performed on an isolated instance in a cloud computing platform with 8 CPU cores and 8~GB of reserved memory \citep{Pace2017}.
Inference on the test set is performed every 250 iterations.
This shows that \gls{BSGP} converges considerably faster in wall-clock time,
even though a single gradient step requires slightly more time.

\begin{figure}[t!]
    \pgfplotsset{every axis title/.append style={yshift=-2ex}}
    \centering\tiny
    \includegraphics{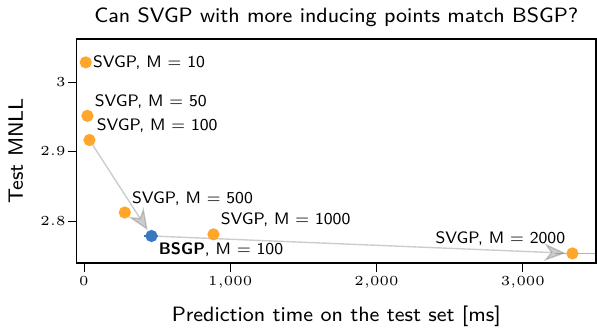}
    \caption{Comparison of test \gls{MNLL} as a function of prediction time on the largest dataset ({Protein}).}
    \label{fig:prediction-timings}
\end{figure}

Computing the predictive distribution, on the other hand, is more challenging as it requires recomputing the covariance matrices $\mathbf{K}_{\mathrm{xz}},\mathbf{K}_\mathrm{zz}$ for each posterior sample $\{\mbZ, \mbtheta\}$, for an overall complexity linear in the number of posterior samples.
This operation can be easily parallelized and implemented on \textsc{gpu}s but it could question the practicality of using a more involved inference method.
In particular, it is relevant to study whether \name{svgp} could deliver superior performance with a higher number of inducing points for less computational overhead.
In \cref{fig:prediction-timings} we study this trade-off on the biggest dataset considered (Protein): while it is evident that predictions with \gls{BSGP} take more time (assuming a serial computation of the covariance matrices), it is also clear that the number of inducing points \name{svgp} requires to (even marginally) improve upon  \gls{BSGP} is significantly larger (up to 20 times).
\setlength\figureheight{.225\textwidth}
\setlength\figurewidth{.4\textwidth}

\begin{figure}[t!]
    \pgfplotsset{every axis title/.append style={yshift=-2ex}}
    \centering\tiny
    \includegraphics{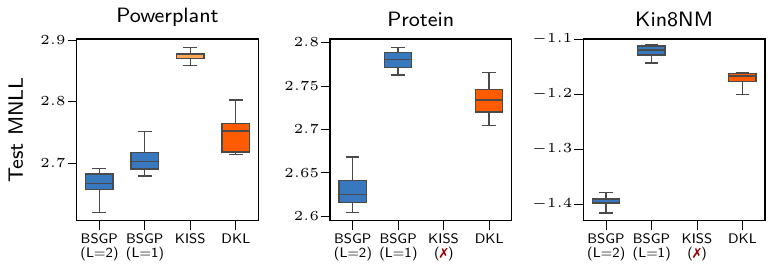}
    \caption{Comparison with structured inducing variables methods. \textsc{kiss-gp} could only run on the Powerplant dataset (hence the \xmark on Protein and Kin8NM).}
    \label{fig:struct-inducing-points}
\end{figure}

\parhead{Structured inducing points.}
Finally, we run one last comparison with methods which exploit structure in the inputs. 
These models allow one to scale the number of inducing variables while maintaining computational tractability. 
\gls{SKI} \citep{Wilson15} proposes to place the inducing inputs on a fixed and equally-spaced grid and to exploit Toeplitz/Kronecker structures with an iterative conjugate gradient method to further enhance scalability.
Despite these benefits, \gls{SKI} is known to fall short with high-dimensional data ($D>4$). This shortcoming was later addressed with \gls{DKL} \citep{Wilson16b}: using a deep neural network \gls{DKL} projects the data in a lower dimensional manifold by learning an useful feature representation, which is then used as input to a \gls{SKI}.
In \cref{fig:struct-inducing-points} we have the comparison of \gls{BSGP} with these two methods.
\gls{SKI} could only run on Powerplant, with a 4-dimensional grid of size 10 (for a total of 10,000 inducing points).
Here, \gls{BSGP} delivers better performance despite having less inducing points.
For \gls{DKL} we followed the suggestion of \citet{Wilson16b} to use a fully-connected neural network with a $[d-1000-1000-500-50-2]$ architecture as feature extractor and a grid size of 100 (for again a total of 10,000 inducing points).
Training is performed by alternating optimization of the neural network weights and the \gls{SKI} parameters.
Thanks to the flexibility of the feature extractor, this configuration is very competitive with our shallow \gls{BSGP}, but it yields lower performance compared to a 2-layer \gls{DGP}.

\setlength\figureheight{.2\textwidth}
\setlength\figurewidth{.2\textwidth}

\subsection{Large scale classification}
The \textsc{airline} dataset is a classic benchmark for large scale classification.
It collects delay information of all commercial flights in USA during 2008, counting more than 5 millions data points.
The goal is to predict if a flight will be delayed based on 8 features, namely {\it month}, {\it day of month}, {\it day of week}, {\it airtime}, {\it distance}, {\it arrival time}, {\it departure time} and {\it age} of the plane.
We pre-process the dataset following the guidelines provided in \citep{Hensman15, Wilson16b}.

After a burn-in phase of 10,000 iterations, we draw 200 samples with 1000 simulation steps in between. We test on 100,000 randomly selected held-out points. We fit three models with $M=100$ inducing points.
\autoref{table:airline} shows the predictive performance of three shallow GP models.
The \gls{BSGP} yields the best test error, \gls{MNLL}, and test \gls{AUC}. We assess the convergence of the predictive posterior by evaluating the $\hat R$-statistics \citep{Gelman2004} over four independent \gls{SGHMC} chains.
This diagnostic yielded a $\hat R = 1.02 \pm 0.045$, which indicates good convergence.
We report further convergence analysis in the supplement. \begin{table}[H]
    \footnotesize
    \caption{\textsc{airline} dataset predictive test performance.}
    \label{table:airline}
    \centering
    \begin{tabular}{lrrr}
        \toprule
        \textbf{Model}     & Error ($\downarrow$) & \textsc{mnll} ($\downarrow$) & \textsc{auc} ($\uparrow$) \\
        \midrule
        \textsc{sghmc-gp}  & $35.85 \% $          & $0.646$                      & $0.671$                   \\
        \textsc{svgp     } & $31.26 \%$           & $0.595$                      & $0.730$                   \\
        \textsc{bsgp     } & $\textbf{30.46} \%$  & $ \textbf{0.580}$            & $\textbf{0.749}$          \\
        \bottomrule
    \end{tabular}
    \hfill
\end{table}

As a further large scale example, we use the \textsc{higgs} dataset \citep{Baldi2014}, which has 11 millions data points with 28 features.
This dataset was created by Monte Carlo simulations of particle dynamics in accelerators to detect the Higgs boson.
We select 90\% of the these points for training, while the rest is kept for testing.
\autoref{table:higgs} reports the final test performance, showing that \gls{BSGP} outperforms the competing methods.
Interestingly, in both these large scale experiments, \textsc{sghmc-gp} always falls back considerably w.r.t. \gls{BSGP} and even \textsc{svgp}.
We argue that, with these large sized datasets, the continuous alternation of optimization of $\mbZ$ and $\mbtheta$ and sampling of $\mbu$ used by the authors (called Moving Window \textsc{mcem}, see \citet{Havasi2018} for  details) might have led to suboptimal solutions.

\begin{table}[t]
    \footnotesize
    \centering
    \captionof{table}{\textsc{higgs} dataset predictive test performance.}
    \label{table:higgs}
    \begin{tabular}{lrrr}
        \toprule
        \textbf{Model}    & Error ($\downarrow$) & \textsc{mnll} ($\downarrow$) & \textsc{auc} ($\uparrow$) \\
        \midrule
        \textsc{sghmc-gp} & $35.39\%$            & $0.628$                      & $0.698$                   \\
        \textsc{svgp    } & $27.79\%$            & $0.544$                      & $0.796$                   \\
        \textsc{bsgp    } & $\mathbf{26.97}\%$   & $\mathbf{0.530}$             & $\mathbf{0.808}$          \\
        \bottomrule
    \end{tabular}
\end{table}

\section{CONCLUSION \& DISCUSSION}

We have developed a fully Bayesian treatment of sparse Gaussian process models that considers the inducing inputs, along with the inducing variables and covariance hyper-parameters, as random variables, places suitable priors and carries out approximate inference over them. 
Our approach, based on \gls{SGHMC}, investigated two conventional priors (Gaussian and uniform) for the inducing inputs as well as two point process based priors (the Determinantal and the Strauss processes). 

By challenging the standard belief of most previous work on sparse \gls{GP} inference that assumes the inducing inputs can be estimated point-wisely, we have developed a state-of-the-art inference method and have demonstrated its outstanding performance on both accuracy and running time on regression and classification problems. 
We hope this work can have an impact similar (or better) to other works in machine learning that have adopted more elaborate Bayesian machinery \citep[e.g.][]{wallach-lda} for long-standing inference problems in commonly used probabilistic models.

Finally, we believe it is worth investigating further more structured priors similar to those presented here (e.g.~exploring different hyper-parameter settings), including a full joint treatment of inducing inputs and their number, i.e.~$p(\mbZ, M)$. 
We leave this for future work.
We are currently investigating ways to extend \gls{BSGP} to convolutional Gaussian process \citep{VanDerWilk17, Dutordoir2019, Blomqvist2018}.

\parhead{Acknowledgements.}
MF gratefully acknowledges support from the AXA Research Fund and the Agence Nationale de la Recherche (grant ANR-18-CE46-0002).

\bibliographystyle{apalike}
\interlinepenalty=100000

\appendix
\onecolumn
\section{A Primer on Inference in Sparse GP Models}
A \gls{GP} defines a distribution over functions $f(\mbx):\bbR^D\rightarrow\bbR$ such that for any subset of points $\{\mbx_1, \ldots, \mbx_N\}$ the function values $\{f(\mbx_1), \ldots, f(\mbx_N)\}$ follow a Gaussian distribution \citep{Rasmussen2005}.
A \gls{GP} is fully described by a mean function $m(\mbx)$ and a covariance function $\kappa(\mbx, \mbx^\prime; \mbtheta)$ with hyper-parameters $\mbtheta$.
Given  a supervised learning problem with $N$ pairs of inputs $\mbx_i$ and labels $y_i$, $\cD = \{(\mbx_i, y_i) | \mbx_i\in\bbR^D, y_i\in\bbR\}_{i=1,\dots,N}$, we consider a \gls{GP}  prior over functions which are fed to a suitable likelihood function to model the observed labels. 

Denoting by $\mbf \in \bbR^N$ the realizations of the \gls{GP} random variables at the $N$ inputs $\mbX = \{\mbx_1, \ldots, \mbx_N\}$ and assuming a zero-mean \gls{GP} prior, we have that $p(\mbf) = \cN(\mathbold{0}, \mbK_{\mathrm{xx}|\mbtheta})$, where $\mbK_{\mathrm{xx}|\mbtheta}$ is the covariance matrix obtained by evaluating $\kappa(\mbx, \mbx^\prime; \mbtheta)$ over all input pairs $\mbx_i, \mbx_j$ (we will drop the explicit parameterization on $\mbtheta$ to keep the notation uncluttered). In the Bayesian setting, given a suitable likelihood function $p(\mby|\mbf)$, the objective is to infer the posterior $p(\mbf|\mby)$ given $N$ pairs of inputs and labels.
This inference problem is analytically tractable for few cases, e.g. using a Gaussian likelihood $\mby|\mbf \sim \cN(\mbf,\sigma^2\mbI)$, but it involves the costly $\bigO(N^3)$ inversion of the covariance matrix $\mbK_\mathrm{xx}$.

Sparse \glspl{GP} are a family of approximate models that address the scalability issue by introducing a set of $M$ inducing variables $\mbu = (u_1, \ldots, u_M)$ at corresponding inducing inputs $\mbZ = \{\mbz_1, \ldots, \mbz_M\}$ such that $u = f(\mbz)$~\citep{Snelson05}.
These inducing variables are assumed to be drawn from the same \gls{GP} as the original process, yielding  the joint prior $p(\mbf,\mbu) = p(\mbu) p(\mbf|\mbu)$ with
\begin{equation}
\label{eq:augmented-prior}
\begin{split}
	p(\mbu)  &= \cN(0, \mbK_\mathrm{zz}) \\
    p(\mbf|\mbu) &= \cN\left(\mbK_\mathrm{xz}\mbK_\mathrm{zz}^{-1}\mbu, \mbK_\mathrm{xx} - \mbK_\mathrm{xz}\mbK_\mathrm{zz}^{-1}\mbK_\mathrm{zx}\right),\,
\end{split}
\end{equation}
where $\mbK_\mathrm{zz} \equiv k(\mbZ, \mbZ)$, $\mbK_{\mathrm{xz}} \equiv k(\mbX, \mbZ)$ and $\mbK_{\mathrm{xz}} = \mbK_{\mathrm{zx}}^T$. 
After introducing the inducing variables, the interest is in obtaining a posterior distribution over $\mbf$ by relying on the set of inducing variables $\mbu$ so as to avoid costly algebraic operations with $\mbK_\mathrm{xx} \in \bbR^{N \times N}$.
A general framework to do this for any likelihood and at scale (using mini-batches) can be obtained using variational inference techniques \cite{Titsias09,hensman13big, bonilla-jmlr-2019}.	 
The main innovation in \citet{Titsias09} is the formulation of an approximate  posterior $q(\mbf,\mbu)$ within  variational inference~\citep{Jordan99}  so as to develop such  a framework. This variational distribution formulation has come to be known as Titsias' trick and has the form:
\begin{equation}
\label{eq:var-posterior}
	q(\mbf,\mbu) = q(\mbu) p(\mbf | \mbu). 
\end{equation}
Following the variational inference approach, and using the above approximate posterior, we introduce the \gls{ELBO},
\begin{equation} \label{eq:elbo}
    \log p(\mathbf{y}) \ge -  \KL{q(\mbu)}{p(\mbu)} + \E_{q(\mbf, \mbu)} \log p(\mby \g \mbf) ,
\end{equation}
where the \gls{KL} term only involves $M$-dimensional distributions, as the conditional prior (\cref{eq:augmented-prior}) is also used in the approximate posterior (\cref{eq:var-posterior}),
which results in the \gls{KL} involving $N$-dimensional distributions vanish.
The second term in the expression above is usually referred to as the \gls{ELL}  and, for factorized conditional likelihoods, it  can be computed efficiently using quadrature or \gls{MC} sampling \citep[see, e.g.,][]{Hensman15b}. 
Thus, posterior estimation under this framework involves constraining $q(\mbu)$ to have a parametric form (usually a Gaussian) and finding its parameters so as to optimize the \gls{ELBO} above. This optimization can be carried out using stochastic-gradient methods operating on mini-batches  yielding a time complexity of $O(M^3)$.

\subsection{MCMC for Variationally Sparse GPs}
\label{sec:mcmc}
An alternative treatment of the inducing variables under the variational framework described above is to avoid constraining $q(\mbu)$ to having any parametric form or admitting simplistic factorizing assumptions. 
As shown by \cite{Hensman15}, this can be, in fact, achieved by finding the optimal (unconstrained) distribution $q(\mbu)$ that maximizes the \gls{ELBO} in \cref{eq:elbo} and sampling from it using techniques such as \gls{MCMC}. 
This optimal distribution can be shown to have the form
\begin{equation}
    \log q(\mbu) = \E_{p(\mbf | \mbu)} \log p(\mby | \mbf)  + \log p(\mbu) + C,
\end{equation}
where $C$ is an unknown normalizing constant.
This expression makes it apparent that sparse variational \gps can be seen as \gp models with a Gaussian prior over the inducing variables and a likelihood which has a complicated form due to the expectation under the conditional $p(\mbf | \mbu)$.
This observation makes it possible to derive \gls{MCMC} samplers for the posterior over $\mbu$, thus relaxing the constraint of having to deal with a fixed form approximation.
The only difficulty is that the likelihood requires the computation of an expectation; however, as mentioned above, for most modeling problems where the likelihood factorizes, this expectation can be calculated as a sum of univariate integrals, for which it is easy to employ numerical quadrature.
\cite{Hensman15} also include the sampling of the hyper-parameters $\mbtheta$ jointly with $\mbu$; however, in order to do this efficiently, a whitening representation is employed, whereby the inducing variables are reparameterized as $\mbu = \mbL_{\mathrm{zz}} \mbnu$, with $\mbK_{\mathrm{zz}} = \mbL_{\mathrm{zz}} \mbL_{\mathrm{zz}}^{\top}$.
The sampling scheme then amounts to sampling from the joint posterior over $\mbnu, \mbtheta$.

The sampling scheme used by \cite{Hensman15} employs a more efficient method based on \acrlong{HMC} \citep[\acrshort{HMC},][]{Duane1987, Neal2010}.
Given a potential energy function defined as $U(\mbu) = -\log p(\mbu, \mby) = -\log p(\mbu|\mby) + C$, \gls{HMC} introduces auxiliary momentum variables $\mbr$ and it generates samples from the joint distribution $p(\mbu,\mbr)$ by simulation of the Hamiltonian dynamics
\begin{align}
    \mathrm{d}\mbu & = \mbM^{-1}\mbr\mathrm{d}t\,,   \nonumber\\
    \mathrm{d}\mbr & = -\nabla U(\mbu)\mathrm{d}t\,, \nonumber
\end{align}
where $\mbM$ is the so called mass matrix, followed by a Metropolis accept/reject step.

\subsection{Stochastic gradient HMC for Deep models}
Different from classic \gls{HMC} where it is required to compute the full gradients  $\nabla U(\mbu) = -\nabla \log p(\mbu|\mby)$, \gls{SGHMC} \citep{Cheni2014} allows to sample from the true intractable posterior by means of stochastic gradients, and without the need of Metropolis accept/reject steps, which would require access to the whole data set.
By modeling the stochastic gradient noise as normally distributed $\cN(\mbzero, \mbV)$, the (discretized) Hamiltonian dynamics are updated as follows
\begin{align}
    \mbu_{t+1} & = \mbu_{t+1} + \varepsilon\mbM^{-1}\mbr_{t}\,,                                                                                 \nonumber\\
    \mbr_{t+1} & = \mbr_t - \varepsilon\widetilde{\nabla U}(\mbu) - \varepsilon\mbC\mbM^{-1}\mbr_{t} + \cN(0, 2\varepsilon(\mbC-\tilde\mbB))\,,\nonumber
\end{align}
where $\varepsilon$ is the step size, $\mbC$ is a user defined friction term and $\tilde\mbB$ is the estimated diffusion matrix of the gradient noise; see e.g., 
\citet{Springenberg2016} for ideas on how to estimate these parameters.

\gls{SGHMC} is the primary inference method used by \citet{Havasi2018} for obtaining samples from the posterior distribution over the latent variables.
Recently, this has been approached using adversarial inference methods \citep{Haibin2019}.

\subsection{Other Approaches to Scalable and Bayesian GPs}
It is worth mentioning that other approaches to scalable inference in \glspl{GP} have been proposed, which feature the possibility to operate using mini-batches. 
For example, looking at the feature-space view of kernel machines, \cite{Rahimi08} show how random features can be obtained for shift invariant covariance functions, like the commonly used squared exponential. These approximations are also useful for addressing the scalability of \glspl{GP} and \glspl{DGP}, as showed by \cite{Gredilla10} and \cite{Cutajar17}.
Similarly, the work on structured approximations of \gps~\cite{Saatci11} has found applications to develop a scalable framework for \gps, later developed to include the possibility to learn deep learning-based representations for the input~\cite{Wilson16}.

The \gls{GPLVM} proposed by \cite{Lawrence05} is a popular approach to Bayesian nonlinear dimensionality reduction and its Bayesian extensions such as those develped by \cite{titsias2010bayesian} consider a prior over the inputs of a \gls{GP}.  Although these methods can be used for training \glspl{GP} with missing or uncertain inputs, we are not aware of previous work adopting such methodologies for inducing inputs within scalable sparse \gls{GP} models.

\section{Discussion on the objectives: VFE vs FITC}
To understand why the \gls{FITC} objective makes sense we need to go back to the original work of \citet{Titsias09,titsias2009techr} and the seminal work of \citet{quinonero2005unifying}. For this, we will consider the regression case and then we can easily generalize our reasoning to the classification case.
\cite{titsias2009techr} shows that, in the standard regression case with isotropic observation noise, his \gls{VFE} optimization framework yields exactly the same predictive posterior as the \gls{PP} approximation \citep{seeger2003fast}, which is referred to as the \gls{DTC} approximation in \cite{quinonero2005unifying}. The optimal variational posterior distribution is given by:
\begin{equation}
    \begin{split}
        \label{eq:optimal-q}
        q^{*}(\mbu \g \mbtheta) &= \Normal(\mbu; \mbm, \mbS), \\
        \mbm &= \sigma^{-2} \Kzz \mbSigma \Kzx \mby \\
        \mbS &= \Kzz \mbSigma \Kzz, \text{ where } \\
        \mbSigma &= \left( \Kzz + \sigma^{-2} \Kzx \Kxz \right)^{-1},
    \end{split}
\end{equation}
where $\sigma^2$ is the observation-noise variance. It is easy to show that, given a Gaussian posterior over the inducing variables with mean and covariance $\mbm$ and $\mbS$, the posterior predictive distribution at test point $\xstar$ is a Gaussian with mean and variance
\begin{equation}
    \begin{split}
        \label{eq:predictive}
        \mu_y(\mbx_\star) &= \kernel(\mbx_\star, \mbZ) \Kzzinv \mbm \\
        \sigma_y^2(\mbx_\star) &= \kernel(\mbx_\star, \mbx_\star) -
        \kernel(\mbx_\star, \mbZ) \Kzzinv \kernel(\mbZ, \mbx_\star) + \kernel(\mbx_\star, \mbZ) \Kzzinv \mbS \Kzzinv \kernel(\mbZ, \mbx_\star).
    \end{split}
\end{equation}
Thus, replacing \cref{eq:optimal-q} in \cref{eq:predictive} we obtain:
\begin{equation}
    \label{eq:predictive-full-vfe}
    \begin{split}
        \mu_y(\mbx_\star) &= \sigma^{-2} \kernel(\mbx_\star, \mbZ) \mbSigma \Kzx \mby \\
        \sigma_y^2(\mbx_\star) &= \kernel(\mbx_\star, \mbx_\star) -
        \kernel(\mbx_\star, \mbZ) \Kzzinv \kernel(\mbZ, \mbx_\star) + \kernel(\mbx_\star, \mbZ) \mbSigma \kernel(\mbZ, \mbx_\star),
    \end{split}
\end{equation}
which indeed corresponds to the predictive distribution of the \gls{DTC}/\gls{PP} approximation. Despite this equivalence, as highlighted in \cite{Titsias09}, the main difference is that the \gls{VFE} framework provides a more robust approach to hyper-parameter estimation as the resulting \gls{ELBO} corresponds to a regularized marginal likelihood of the \gls{DTC} approach and hence should be more robust to overfitting. Nevertheless, the \gls{DTC}/\gls{PP}, and consequently the \gls{VFE}, predictive distribution has been shown to be less accurate than the \gls{FITC} approximation \citep{Titsias09,quinonero2005unifying,snelson2007flexible}.
\subsection{The FITC Approximation}
The \gls{FITC} approximation  considers the following approximate conditional prior:
\begin{align}
    p(\mbf \g \mbu)    & \approx \Normal\left(\mbf;  \Kxz \Kzzinv \mbu, \diag\left(\Kxx - \Kxz \Kzzinv \Kzx \right) \right) \nonumber\\
                       & = \prod_{n=1}^N p(f_n \g \mbu)                                                                     
                        = \prod_{n=1}^N \Normal(f_n; \tilde{\mu}_n, \tilde{\sigma}_n^2), \text{ with }                     \nonumber\\
    \label{fitc-mean-n}
    \tilde{\mu}_n      & = \kernel(\mbx_n,\mbZ) \Kzzinv \mbu                                                                \\
    \label{fitc-var-n}
    \tilde{\sigma}_n^2 & = \kernel(\mbx_n, \mbx_n) - \kernel(\mbx_n, \mbZ) \Kzzinv \kernel(\mbZ, \mbx_n) \text{.}
\end{align}
As we shall see later, is this factorization assumption in the conditional prior that will yield a decomposable objective amenable to stochastic gradient techniques. For now, consider the posterior predictive distribution under the \gls{FITC} approximation\footnote{Which is, in fact, the same as in the \gls{SPGP} framework of \citet{Snelson07}.}
\begin{equation}
    \label{eq:predictive-full-fitc}
    \begin{split}
        \mu_{\gls{FITC}}(\xstar) &= \kernel(\xstar, \mbZ) \mbSigma_{\gls{FITC}} \Kzx \mbLambda^{-1} \mby\\
        \sigma^2_{\gls{FITC}}(\xstar) &= \kernel(\xstar, \xstar) - \kernel(\xstar, \mbZ) \Kzzinv \kernel(\mbZ, \xstar) + \kernel(\xstar, \mbZ) \mbSigma_{\gls{FITC}} \kernel(\mbZ, \xstar), \text{ where }\\
        \mbLambda &= \diag(\Kxx - \Kxz \Kzzinv \Kzx + \sigma^2 \mbI) \text{ and } \\
        \mbSigma_{\gls{FITC}} &= (\Kzz + \Kzx \mbLambda^{-1} \Kxz)^{-1}.
    \end{split}
\end{equation}
We now see why \gls{FITC}'s predictive distribution above is more accurate than \gls{VFE}'s in \cref{eq:predictive-full-vfe}, as we can obtain \gls{FITC}'s  by replacing $\sigma^2 \mbI$ in \gls{VFE}'s solution with $\diag(\Kxx - \Kxz \Kzzinv \Kzx) + \sigma^2 \mbI$. Effectively, as described in \cite{quinonero2005unifying}, \gls{VFE}'s solution (which is the same as \gls{DTC}'s) can be understood as considering a deterministic conditional prior $p(\mbf \g \mbu)$, i.e.~with zero variance.
\subsection{Stochastic Updates Using the FITC Approximation}
Now we can understand why the log of the expectation can provide more accurate results than the expectation of the log. Basically in the former we are using the \gls{FITC} approximation while in the later we are using the \gls{VFE}/\gls{DTC}/\gls{PP} approximation. It is easy to show that when using the \gls{FITC} approximation, one can obtain a decomposable objective function that can be implemented at large scale using stochastic gradient techniques. Here we focus only on the expectation of the conditional likelihood (which is the crucial term) and in the regression setting for simplicity but the extension to the classification case (e.g. using quadrature) is straightforward.
\begin{align}
    \log p(\mby, \mbu \g \mbtheta) & = \log \E_{p(\mbf \g \mbu, \mbtheta)} \left[ p(\mby \g \mbf \right)]                                                          \nonumber\\
                                   & = \log \int_{\mbf} p(\mbf \g \mbu, \mbtheta ) p(\mby \g \mbf) \d\mbf  {\text{\tiny \quad will drop $\mbtheta$ for simplicity from now on}} \nonumber\\
                                   & = \log \int_{f_1, \dots, f_N} \prod_{n=1}^N p(y_n \g f_n) p(f_n \g \mbu) \d{\mbf}                                               \nonumber\\
                                   & = \log  \prod_{n=1}^N \int_{f_n} \Normal(y_n; f_n, \sigma^2) \Normal(f_n ; \tilde{\mu}_n, \tilde{\sigma}^2_n) \d{f_n}           \nonumber\\
                                   & = \log \prod_{n=1}^N p(y_n \g \mbu)                                                                                           \nonumber\\
                                   & = \sum_{n=1}^N \log \Normal(y_n; \tilde{\mu}_n, \tilde{\sigma}_n^2 + \sigma^2),
\end{align}
where $\tilde{\mu}_n, \tilde{\sigma}_n^2$ are given by \cref{fitc-mean-n} and \cref{fitc-var-n}.
\paragraph{Binary classification.}
Similar results can be derived  for binary classification with Bernoulli likelihood and response function $\lambda(f)$: 
\begin{equation}
\label{eq:E-class}
\log p(\mby, \mbu \g \mbtheta) = \log \E_{p(\mbf \g \mbu, \mbtheta)} \left[ p(\mby \g \mbf \right)] = \log \prod_{n=1}^N \int_{f_n} \Normal(f_n ; \tilde{\mu}_n, \tilde{\sigma}^2_n) \text{Bern}(y_n; \lambda(f_n)) \d{f_n} .
\end{equation}
When the response function is the cdf of a standard Normal distribution, i.e., $\lambda(f_n) = \Phi(f_n) \defeq \int_{- \infty}^{f_n} \Normal(f_n; 0, 1) d f_n$, which is also known as the probit regression model, the expectation above can be computed analytically to obtain:
\begin{equation}
    \log p(\mby, \mbu \g \mbtheta) = \sum_{n=1}^N \log \text{Bern}(y_n; \Phi(\tilde{\mu}_n / \sqrt{1 + \tilde{\sigma}_n^2})).
\end{equation}
For other response functions the expectation in \cref{eq:E-class} can be estimated using quadrature.

\subsection{An heteroskedastic version of the Gaussian likelihood}
As \citet{titsias2009techr} discussed in Appendix C, the \gls{FITC} approximation corresponds to a \gls{GP} regression with heteroskedastic noise variance
\begin{align}
    p(\mby | \mbf) & = \mathcal{N}(\mby | \mbf, \sigma^2\mbI + \diag[\mbK_{\mbx\mbx} - \mbK_{\mbx\mbz}\mbK_{\mbz\mbz}\mbK_{\mbz\mbx}]).
\end{align}
If we apply this augmented likelihood to the variational expectations term, we get
\begin{align}
    \mathbb{E}_{q(\mbf)}\log p(\mby|\mbf, \sigma^2, \mbtheta) & = -\frac{1}{2} \sum_{j=1}^n\left(\log 2\pi (\sigma^2 + \tilde{\sigma}_j^2) + \frac{(y_j-\tilde{\mu}_j)^2+\tilde{\sigma}_j^2}{\sigma^2 + \tilde{\sigma}_j^2}\right).
\end{align}
Since \cite{titsias2009techr} considers this \gls{VFE} formulation, we also compare with it.

\subsection{Concluding remarks}
The main reasoning in \cite{Titsias09}'s work behind the better performance of \gls{VFE}, despite providing a less accurate predictive posterior than \gls{FITC}'s, was that hyper-parameter estimation was more robust due to the use of the variational objective (which provided an extra regularization term). 
Now, we have a better way to do inference on hyper-parameters and inducing inputs by placing priors on those and by carrying out free-form inference upon them   with \gls{SGHMC}.

\section{Extension to Deep Gaussian Processes}
In this section, we derive the mathematical basis for  of Bayesian treatment of inducing inputs in a \gls{DGP} setting \citep{Damianou13}. 
We assume a deep Gaussian process prior $f^L\circ f^{L-1}\circ\cdots f^1$, where each $f^l$ is a \gp. 
For notational brevity, we use $\mbtheta^l$ as both kernel hyper-parameters and inducing inputs of the $l$-th layer, and $\mbf^0$ as the input vector $\mbX$. 
Then we can write down the joint distribution over visible and latent variables (omitting the dependency on $\mbX$ for clarity) as
\begin{equation}
    p\left(\mby, \left\{\mbf^l, \mbu^l, \mbtheta^l\right\}_{l=1}^{L} \right) =  p\left(\mby   \given  \mbf^L \right) \prod_{l=1}^{L} p\left(\mbf^l  \given  \mbu^l, \mbf^{l-1}, \mbtheta^{l} \right)p\left(\mbu^l \given \mbtheta^l\right)p\left(\mbtheta^l\right). \label{eq:joint_lik}
\end{equation}
Our goal is to estimate the posterior, \begin{align}
   & \log \widetilde{p}\left(\left\{\mbu^l, \mbtheta^l\right\}_{l=1}^{L} \given \mby\right) = \nonumber\\
   & \quad = \log\E_{p\left(\left\{\mbf^l\right\} \given \left\{\mbu^l, \mbtheta^l\right\}\right)} p\left(\mby  \given  \mbf^L\right) + \sum_{l=1}^L \left(\log p\left(\mbu^l \given \mbtheta^l\right) + \log p\left(\mbtheta^l\right)\right) - \log C. \label{eq:optim}
    \end{align}
Here $C$ is a normalizing constant, after integrating out ${\{\mbf^l, \mbu^l, \mbtheta^l\}_{l=1}^{L}}$ from the joint.
While the distribution $\widetilde{p}$ is intractable, we have obtained the form of its (un-normalized) log joint, from which we can sample using \gls{HMC} methods. However, \cref{eq:optim} is not immediately computable owing to the intractable expectation term. 
More calculations reveal that we can, nevertheless, obtain estimates of this expectation term with Monte Carlo sampling
\begin{align}
    & \log \E_{p\left(\left\{\mbf^l\right\}  \given \left\{\mbu^l, \mbtheta^l\right\}\right)} p\left(\mby\given\mbf^L\right)  \approx \nonumber\\
    & \qquad \approx \log\E_{p\left(\left\{\mbf^l\right\}_{l=2}^{L}\given\widetilde{\mbf}^1, \left\{\mbu^l, \mbtheta^l\right\}_{l=2}^L\right)}  p\left(\mby\given\mbf^L\right) , \quad \widetilde{\mbf}^1\sim p\left(\mbf^1\given\mbu^1, \mbtheta^1, \mbf^0\right), \nonumber\\
    &  \qquad \approx \log\E_{p\left(\left\{\mbf^l\right\}_{l=3}^{L}\given\widetilde{\mbf}^2,\left\{\mbu^l, \mbtheta^l\right\}_{l=3}^L\right)}  p\left(\mby\given\mbf^L\right) , \quad \widetilde{\mbf}^2\sim p\left(\mbf^2\given\mbu^2, \mbtheta^2, \widetilde{\mbf}^1\right),\nonumber\\
    & \qquad \approx \hdots \nonumber\\
    & \qquad \approx \log\E_{p\left(\mbf^L\given\widetilde{\mbf}^{L-1}, \mbu^{L}, \mbtheta^L\right)} p\left(\mby\given\mbf^L\right),\quad \widetilde{\mbf}^{L-1} \sim p\left(\mbf^{L-1}\left\vert\mbu^{L-1}, \mbtheta^{L-1}, \widetilde{\mbf}^{L-2}\right.\right),\nonumber\\
    & \qquad = \sum_{n=1}^N \log\E_{p\left(f_n^L\given\widetilde{f}_n^{L-1}, \mbu^{L}, \mbtheta^L\right)} p\left(y_n\given f_n^L\right)
    \end{align}
Because of the layer-wise factorization of the joint likelihood (\cref{eq:joint_lik}), each step of the approximation is unbiased. 
While it is possible to approximate the last-layer expectation with a Monte Carlo sample $\widetilde{f}_j^L$, the expectation is tractable when $y_j\g f_j^L$ is a Gaussian or a Bernoulli distribution with a probit regression model, or is computable with one-dimensional quadrature \citep{Hensman15}.

\clearpage 
\section{Additional Results}

\setlength\figureheight{.17\textwidth}
\setlength\figurewidth{.27\textwidth}

\begin{figure*}[!h!]
    \begin{minipage}{.49\textwidth}
        \centering\tiny
        \pgfplotsset{every axis title/.append style={yshift=-1ex}}
        \includegraphics{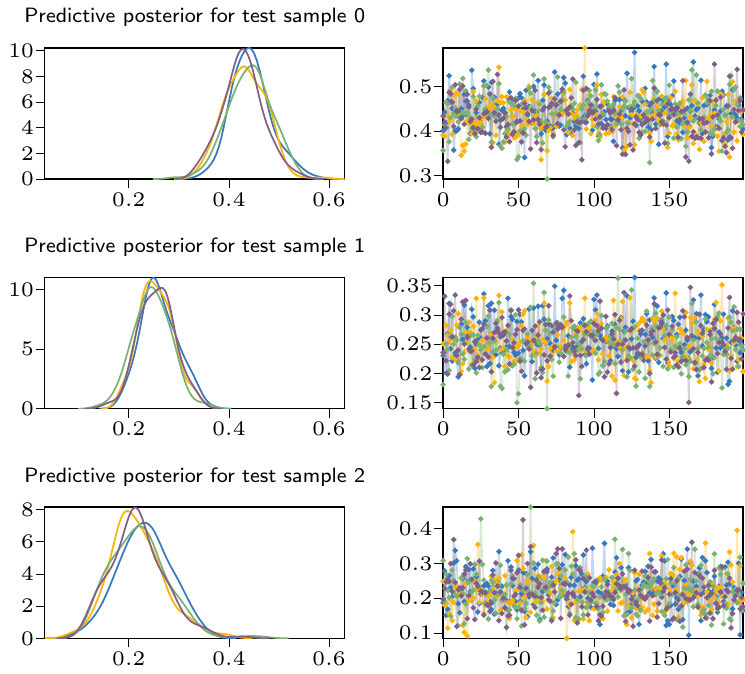}
        \captionof{figure}{Traces for three test points on the Airline dataset (4 chains/200 samples).}
        \label{fig:trace-airline}
    \end{minipage}
    \begin{minipage}{.49\textwidth}
        \centering\tiny
        \centering
        \captionof{table}{Datasets used, including number of datapoints and their dimensionality.}
        \scriptsize\sc
        \begin{tabular}{r|rrr}
    \toprule
    \textbf{ name } & \textbf{ n. } & \textbf{ d-in } & \textbf{ d-out } \\
    \midrule
    boston          & 506           & 13              & 1                \\
    concrete        & 1,030         & 8               & 1                \\
    energy          & 768           & 8               & 2                \\
    kin8nm          & 8,192         & 8               & 1                \\
    naval           & 11,934        & 16              & 2                \\
    powerplant      & 9,568         & 4               & 1                \\
    protein         & 45,730        & 9               & 1                \\
    yacht           & 308           & 6               & 1                \\
    \midrule
    airline         & 5,934,530     & 8               & 2                \\
    higgs           & 11,000,000    & 28              & 2                \\
    \bottomrule
\end{tabular}
    \end{minipage}
\end{figure*}

\setlength\figureheight{.22\textwidth}
\setlength\figurewidth{.30\textwidth}
\begin{figure}[H]
    \pgfplotsset{every axis title/.append style={yshift=-2ex}}
    \centering\tiny
    \includegraphics{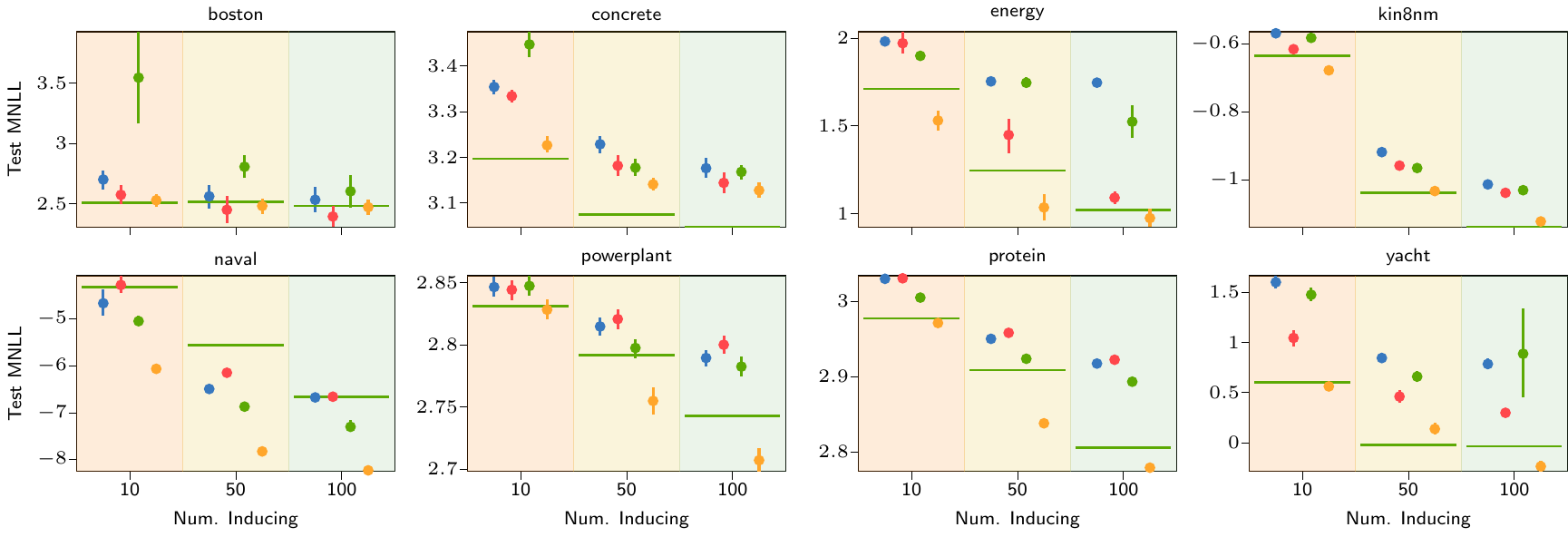}
    \setlength\figureheight{.215\textwidth}
    \setlength\figurewidth{.23\textwidth}
    \includegraphics{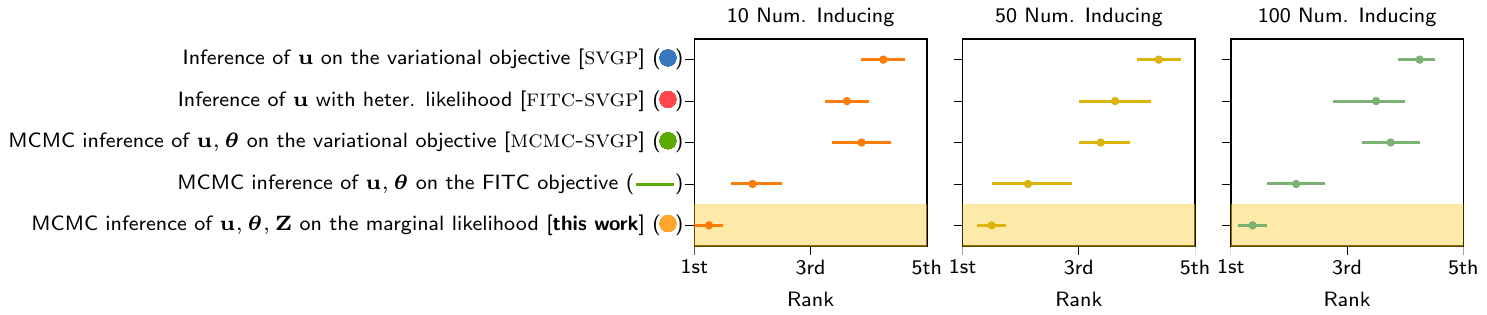}
    \tikzexternaldisable
    \definecolor{color0}{rgb}{0.215686274509804,0.470588235294118,0.749019607843137}
    \definecolor{color1}{rgb}{1,0.27843137254902,0.298039215686275}
    \definecolor{color2}{rgb}{0.36078431372549,0.662745098039216,0.0156862745098039}
    \definecolor{color3}{rgb}{1,0.650980392156863,0.168627450980392}
    \definecolor{color4}{rgb}{0.52156862745098,0.403921568627451,0.596078431372549}
    \caption{Empirical analysis of different choices of objectives for optimization and sampling.
        }
    \tikzexternalenable
    \label{fig:ablation_study_objectives}
\end{figure}

\setlength\figureheight{.22\textwidth}
\setlength\figurewidth{.30\textwidth}
\begin{figure}[t!]
    \pgfplotsset{every axis title/.append style={yshift=-2ex}}
    \centering\tiny
    \includegraphics{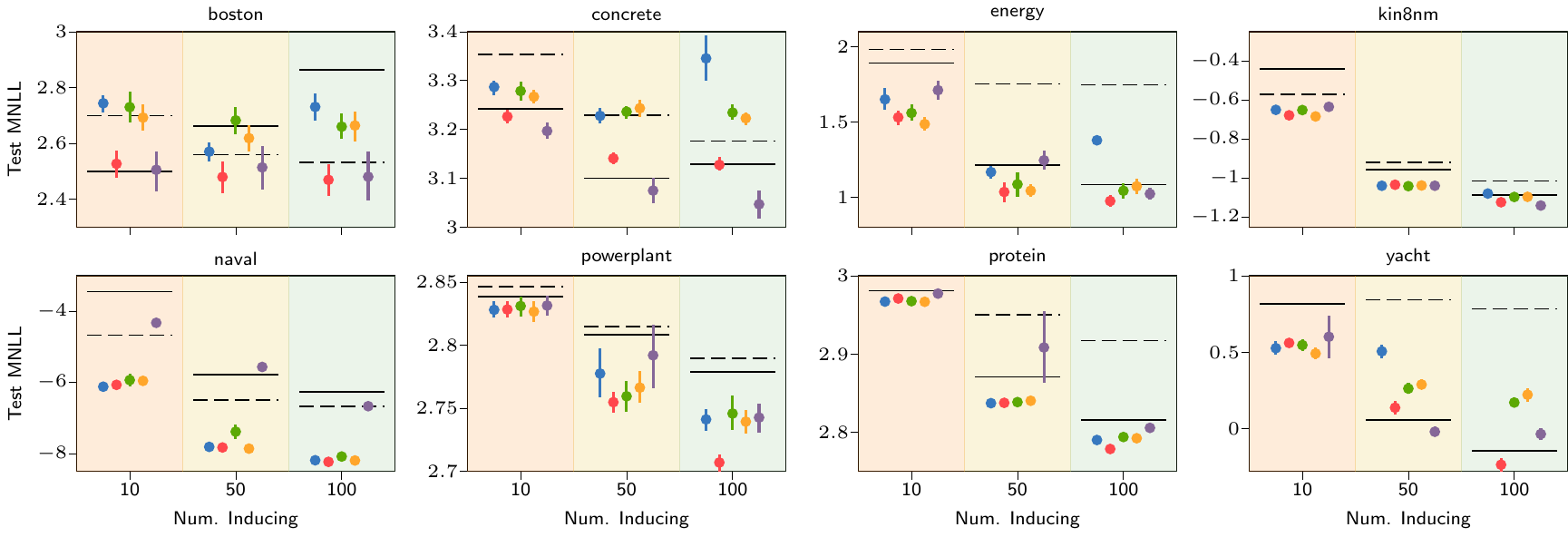}
    \\[4ex]
    \setlength\figureheight{.25\textwidth}
    \setlength\figurewidth{.28\textwidth}
    \includegraphics{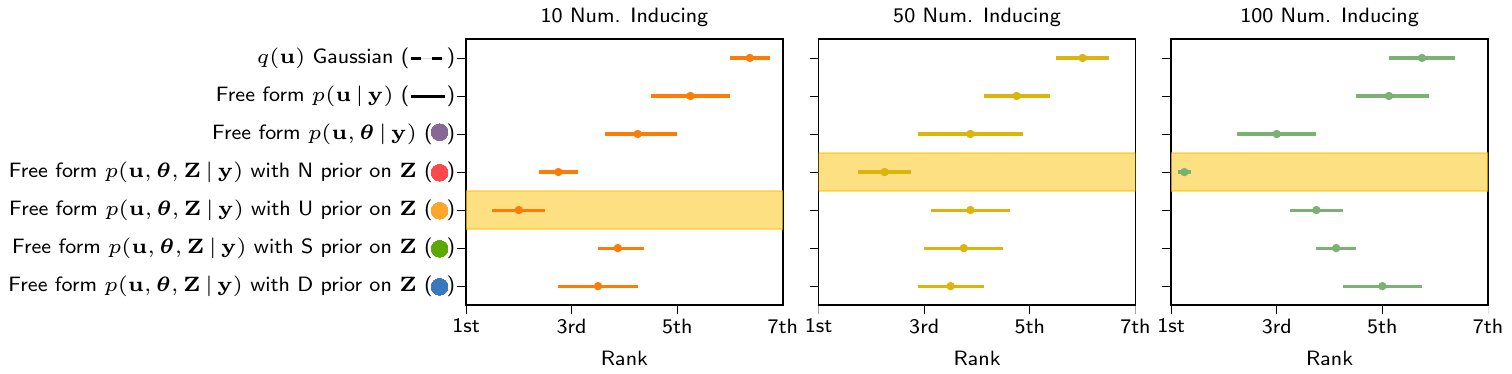}
    \tikzexternaldisable
    \definecolor{color0}{rgb}{0.215686274509804,0.470588235294118,0.749019607843137}
    \definecolor{color1}{rgb}{1,0.27843137254902,0.298039215686275}
    \definecolor{color2}{rgb}{0.36078431372549,0.662745098039216,0.0156862745098039}
    \definecolor{color3}{rgb}{1,0.650980392156863,0.168627450980392}
    \definecolor{color4}{rgb}{0.52156862745098,0.403921568627451,0.596078431372549}
    \caption{Ablation study on the Test \name{mnll} based on the UCI benchmark for different number of inducing variables and for
        determinantal point process prior
        ({\protect\tikz \protect\draw[thick, color=color0, fill=color0] plot[mark=*, line width=5pt, mark options={scale=1.3}] (0,0.5);}),
        normal prior
        ({\protect\tikz \protect\draw[thick, color=color1, fill=color1] plot[mark=*, line width=5pt, mark options={scale=1.3}] (0,0.5);}),
        Strauss process prior
        ({\protect\tikz \protect\draw[thick, color=color2, fill=color2] plot[mark=*, line width=5pt, mark options={scale=1.3}] (0,0.5);}) and
        uniform prior
        ({\protect\tikz \protect\draw[thick, color=color3, fill=color3] plot[mark=*, line width=5pt, mark options={scale=1.3}] (0,0.5);}).
        These are compared with
        ({\protect\tikz \protect\draw[thick, color=color4, fill=color4] plot[mark=*, line width=5pt, mark options={scale=1.3}] (0,0.5);}),
        corresponding to the case of inducing positions optimized and inducing variables and covariance hyper-parameters sampled,
        with
        ({\protect\tikz[baseline=-0.5ex]\protect\draw[thick, color=black, fill=black] plot[] (0,0)--+(.35,0);}) where only inducing variables are inferred, while the rest is optimized (similarly to \textsc{sghmc-dgp}).
        Finally ({\protect\tikz[baseline=-0.5ex]\protect\draw[thick, color=black, fill=black, dashed] plot[] (0,0)--+(.35,0);}) is the classic \textsc{svgp}, where everything is optimized.
    }
    \tikzexternalenable
    \label{fig:ablation_study}
\end{figure}

\begin{table}
    \centering
    \captionof{table}{Tabular version of Figure 7 in the main paper.}
    \scriptsize\sc
    \renewcommand{\tabcolsep}{1.5ex}
\begin{tabular}{l|cccccccc}
\toprule
{} & \multicolumn{8}{c}{test mnll} \\
dataset &                           boston &                         concrete &                           energy &                            kin8nm &                             naval &                       powerplant &                          protein &                             yacht \\
name            &                                  &                                  &                                  &                                   &                                   &                                  &                                  &                                   \\
\midrule
bsgp 1 &  $2.47$ \scalebox{0.8}{($0.16$)}&  $3.12$ \scalebox{0.8}{($0.04$)}&  $0.97$ \scalebox{0.8}{($0.13$)}&  $-1.12$ \scalebox{0.8}{($0.01$)}&  $-8.22$ \scalebox{0.8}{($0.04$)}&  $2.71$ \scalebox{0.8}{($0.02$)}&  $2.78$ \scalebox{0.8}{($0.01$)}&  $-0.23$ \scalebox{0.8}{($0.13$)}\\
bsgp 2 &  $2.47$ \scalebox{0.8}{($0.15$)}&  $3.04$ \scalebox{0.8}{($0.05$)}&  $0.95$ \scalebox{0.8}{($0.16$)}&  $-1.40$ \scalebox{0.8}{($0.01$)}&  $-8.23$ \scalebox{0.8}{($0.04$)}&  $2.67$ \scalebox{0.8}{($0.02$)}&  $2.63$ \scalebox{0.8}{($0.02$)}&  $-0.72$ \scalebox{0.8}{($0.15$)}\\
bsgp 3 &  $2.47$ \scalebox{0.8}{($0.14$)}&  $2.96$ \scalebox{0.8}{($0.10$)}&  $0.95$ \scalebox{0.8}{($0.15$)}&  $-1.41$ \scalebox{0.8}{($0.01$)}&  $-8.02$ \scalebox{0.8}{($0.04$)}&  $2.66$ \scalebox{0.8}{($0.03$)}&  $2.57$ \scalebox{0.8}{($0.03$)}&  $-0.83$ \scalebox{0.8}{($0.10$)}\\
bsgp 4 &  $2.48$ \scalebox{0.8}{($0.14$)}&  $2.97$ \scalebox{0.8}{($0.06$)}&  $0.92$ \scalebox{0.8}{($0.14$)}&  $-1.43$ \scalebox{0.8}{($0.02$)}&  $-8.03$ \scalebox{0.8}{($0.05$)}&  $2.65$ \scalebox{0.8}{($0.05$)}&  $2.50$ \scalebox{0.8}{($0.03$)}&  $-0.76$ \scalebox{0.8}{($0.13$)}\\
bsgp 5 &  $2.48$ \scalebox{0.8}{($0.12$)}&  $2.91$ \scalebox{0.8}{($0.08$)}&  $0.75$ \scalebox{0.8}{($0.30$)}&  $-1.43$ \scalebox{0.8}{($0.01$)}&  $-8.09$ \scalebox{0.8}{($0.05$)}&  $2.65$ \scalebox{0.8}{($0.03$)}&  $2.43$ \scalebox{0.8}{($0.03$)}&  $-0.74$ \scalebox{0.8}{($0.08$)}\\
ipvi gp 1       &  $2.84$ \scalebox{0.8}{($0.36$)}&  $3.19$ \scalebox{0.8}{($0.11$)}&  $1.27$ \scalebox{0.8}{($0.07$)}&  $-1.12$ \scalebox{0.8}{($0.02$)}&  $-5.96$ \scalebox{0.8}{($0.89$)}&  $2.79$ \scalebox{0.8}{($0.03$)}&  $2.81$ \scalebox{0.8}{($0.02$)}&   $1.21$ \scalebox{0.8}{($1.50$)}\\
ipvi gp 2       &  $2.73$ \scalebox{0.8}{($0.35$)}&  $3.13$ \scalebox{0.8}{($0.11$)}&  $1.31$ \scalebox{0.8}{($0.28$)}&  $-1.34$ \scalebox{0.8}{($0.02$)}&  $-4.98$ \scalebox{0.8}{($0.48$)}&  $2.76$ \scalebox{0.8}{($0.07$)}&  $2.65$ \scalebox{0.8}{($0.02$)}&   $0.74$ \scalebox{0.8}{($1.13$)}\\
ipvi gp 3       &  $2.61$ \scalebox{0.8}{($0.25$)}&  $3.08$ \scalebox{0.8}{($0.13$)}&  $1.21$ \scalebox{0.8}{($0.12$)}&  $-1.33$ \scalebox{0.8}{($0.03$)}&  $-4.86$ \scalebox{0.8}{($0.23$)}&  $2.72$ \scalebox{0.8}{($0.06$)}&  $2.74$ \scalebox{0.8}{($0.05$)}&   $1.05$ \scalebox{0.8}{($1.77$)}\\
ipvi gp 4       &  $2.64$ \scalebox{0.8}{($0.44$)}&  $3.11$ \scalebox{0.8}{($0.18$)}&  $1.19$ \scalebox{0.8}{($0.25$)}&  $-1.33$ \scalebox{0.8}{($0.01$)}&  $-4.94$ \scalebox{0.8}{($0.20$)}&  $2.76$ \scalebox{0.8}{($0.02$)}&  $2.79$ \scalebox{0.8}{($0.01$)}&   $2.47$ \scalebox{0.8}{($2.34$)}\\
ipvi gp 5       &  $2.51$ \scalebox{0.8}{($0.20$)}&  $3.08$ \scalebox{0.8}{($0.17$)}&  $1.15$ \scalebox{0.8}{($0.22$)}&  $-1.29$ \scalebox{0.8}{($0.02$)}&  $-5.09$ \scalebox{0.8}{($0.49$)}&  $2.72$ \scalebox{0.8}{($0.04$)}&  $2.80$ \scalebox{0.8}{($0.01$)}&   $2.84$ \scalebox{0.8}{($3.64$)}\\
sghmc gp 1      &  $2.82$ \scalebox{0.8}{($0.33$)}&  $3.13$ \scalebox{0.8}{($0.09$)}&  $1.08$ \scalebox{0.8}{($0.28$)}&  $-1.08$ \scalebox{0.8}{($0.01$)}&  $-6.23$ \scalebox{0.8}{($0.14$)}&  $2.76$ \scalebox{0.8}{($0.05$)}&  $2.81$ \scalebox{0.8}{($0.01$)}&  $-0.11$ \scalebox{0.8}{($0.28$)}\\
sghmc gp 2      &  $2.77$ \scalebox{0.8}{($0.37$)}&  $2.99$ \scalebox{0.8}{($0.07$)}&  $0.91$ \scalebox{0.8}{($0.15$)}&  $-1.32$ \scalebox{0.8}{($0.01$)}&  $-6.57$ \scalebox{0.8}{($0.11$)}&  $2.72$ \scalebox{0.8}{($0.04$)}&  $2.71$ \scalebox{0.8}{($0.02$)}&  $-0.52$ \scalebox{0.8}{($0.14$)}\\
sghmc gp 3      &  $2.78$ \scalebox{0.8}{($0.28$)}&  $3.02$ \scalebox{0.8}{($0.16$)}&  $0.91$ \scalebox{0.8}{($0.14$)}&  $-1.37$ \scalebox{0.8}{($0.02$)}&  $-6.56$ \scalebox{0.8}{($0.09$)}&  $2.68$ \scalebox{0.8}{($0.02$)}&  $2.66$ \scalebox{0.8}{($0.03$)}&  $-0.57$ \scalebox{0.8}{($0.19$)}\\
sghmc gp 4      &  $2.75$ \scalebox{0.8}{($0.34$)}&  $2.98$ \scalebox{0.8}{($0.13$)}&  $0.69$ \scalebox{0.8}{($0.22$)}&  $-1.38$ \scalebox{0.8}{($0.02$)}&  $-6.42$ \scalebox{0.8}{($0.08$)}&  $2.67$ \scalebox{0.8}{($0.04$)}&  $2.62$ \scalebox{0.8}{($0.02$)}&  $-0.69$ \scalebox{0.8}{($0.12$)}\\
sghmc gp 5      &  $3.75$ \scalebox{0.8}{($1.91$)}&  $3.11$ \scalebox{0.8}{($0.21$)}&  $1.00$ \scalebox{0.8}{($0.42$)}&  $-1.39$ \scalebox{0.8}{($0.02$)}&  $-6.55$ \scalebox{0.8}{($0.09$)}&  $2.65$ \scalebox{0.8}{($0.04$)}&  $2.59$ \scalebox{0.8}{($0.02$)}&  $-0.53$ \scalebox{0.8}{($0.18$)}\\
svgp 1          &  $2.53$ \scalebox{0.8}{($0.25$)}&  $3.18$ \scalebox{0.8}{($0.05$)}&  $1.75$ \scalebox{0.8}{($0.06$)}&  $-1.01$ \scalebox{0.8}{($0.01$)}&  $-6.67$ \scalebox{0.8}{($0.09$)}&  $2.79$ \scalebox{0.8}{($0.02$)}&  $2.92$ \scalebox{0.8}{($0.01$)}&   $0.78$ \scalebox{0.8}{($0.13$)}\\
\bottomrule
\end{tabular}

\end{table}
\begin{table}
    \centering
    \captionof{table}{Normalized RMSE  corresponding to results of Figure 7 in the main paper.}
    \scriptsize\sc
    \renewcommand{\tabcolsep}{1.5ex}
\begin{tabular}{l|cccccccc}
\toprule
{} & \multicolumn{8}{c}{test error} \\
dataset &                           boston &                         concrete &                           energy &                           kin8nm &                            naval &                       powerplant &                          protein &                            yacht \\
name            &                                  &                                  &                                  &                                  &                                  &                                  &                                  &                                  \\
\midrule
bsgp 1 &  $0.36$ \scalebox{0.8}{($0.07$)} &  $0.40$ \scalebox{0.8}{($0.03$)} &  $0.13$ \scalebox{0.8}{($0.01$)} &  $0.31$ \scalebox{0.8}{($0.01$)} &  $0.02$ \scalebox{0.8}{($0.00$)} &  $0.23$ \scalebox{0.8}{($0.00$)} &  $0.72$ \scalebox{0.8}{($0.00$)} &  $0.04$ \scalebox{0.8}{($0.01$)} \\
bsgp 2 &  $0.37$ \scalebox{0.8}{($0.07$)} &  $0.36$ \scalebox{0.8}{($0.02$)} &  $0.13$ \scalebox{0.8}{($0.01$)} &  $0.24$ \scalebox{0.8}{($0.00$)} &  $0.01$ \scalebox{0.8}{($0.00$)} &  $0.22$ \scalebox{0.8}{($0.00$)} &  $0.69$ \scalebox{0.8}{($0.00$)} &  $0.03$ \scalebox{0.8}{($0.01$)} \\
bsgp 3 &  $0.37$ \scalebox{0.8}{($0.07$)} &  $0.32$ \scalebox{0.8}{($0.03$)} &  $0.13$ \scalebox{0.8}{($0.01$)} &  $0.24$ \scalebox{0.8}{($0.01$)} &  $0.01$ \scalebox{0.8}{($0.00$)} &  $0.22$ \scalebox{0.8}{($0.00$)} &  $0.68$ \scalebox{0.8}{($0.00$)} &  $0.03$ \scalebox{0.8}{($0.01$)} \\
bsgp 4 &  $0.37$ \scalebox{0.8}{($0.07$)} &  $0.33$ \scalebox{0.8}{($0.02$)} &  $0.13$ \scalebox{0.8}{($0.01$)} &  $0.24$ \scalebox{0.8}{($0.01$)} &  $0.01$ \scalebox{0.8}{($0.00$)} &  $0.21$ \scalebox{0.8}{($0.00$)} &  $0.67$ \scalebox{0.8}{($0.01$)} &  $0.03$ \scalebox{0.8}{($0.01$)} \\
bsgp 5 &  $0.37$ \scalebox{0.8}{($0.07$)} &  $0.32$ \scalebox{0.8}{($0.03$)} &  $0.12$ \scalebox{0.8}{($0.03$)} &  $0.23$ \scalebox{0.8}{($0.00$)} &  $0.01$ \scalebox{0.8}{($0.00$)} &  $0.21$ \scalebox{0.8}{($0.00$)} &  $0.65$ \scalebox{0.8}{($0.01$)} &  $0.03$ \scalebox{0.8}{($0.01$)} \\
ipvi gp 1       &  $0.34$ \scalebox{0.8}{($0.04$)} &  $0.34$ \scalebox{0.8}{($0.02$)} &  $0.13$ \scalebox{0.8}{($0.01$)} &  $0.31$ \scalebox{0.8}{($0.01$)} &  $0.16$ \scalebox{0.8}{($0.17$)} &  $0.23$ \scalebox{0.8}{($0.00$)} &  $0.72$ \scalebox{0.8}{($0.01$)} &  $0.04$ \scalebox{0.8}{($0.02$)} \\
ipvi gp 2       &  $0.35$ \scalebox{0.8}{($0.06$)} &  $0.32$ \scalebox{0.8}{($0.02$)} &  $0.13$ \scalebox{0.8}{($0.01$)} &  $0.25$ \scalebox{0.8}{($0.01$)} &  $0.62$ \scalebox{0.8}{($0.22$)} &  $0.22$ \scalebox{0.8}{($0.01$)} &  $0.68$ \scalebox{0.8}{($0.01$)} &  $0.03$ \scalebox{0.8}{($0.02$)} \\
ipvi gp 3       &  $0.34$ \scalebox{0.8}{($0.05$)} &  $0.30$ \scalebox{0.8}{($0.03$)} &  $0.13$ \scalebox{0.8}{($0.01$)} &  $0.25$ \scalebox{0.8}{($0.01$)} &  $0.65$ \scalebox{0.8}{($0.09$)} &  $0.22$ \scalebox{0.8}{($0.01$)} &  $0.65$ \scalebox{0.8}{($0.01$)} &  $0.03$ \scalebox{0.8}{($0.02$)} \\
ipvi gp 4       &  $0.33$ \scalebox{0.8}{($0.06$)} &  $0.31$ \scalebox{0.8}{($0.03$)} &  $0.12$ \scalebox{0.8}{($0.03$)} &  $0.25$ \scalebox{0.8}{($0.00$)} &  $0.70$ \scalebox{0.8}{($0.01$)} &  $0.22$ \scalebox{0.8}{($0.01$)} &  $0.65$ \scalebox{0.8}{($0.01$)} &  $0.04$ \scalebox{0.8}{($0.03$)} \\
ipvi gp 5       &  $0.32$ \scalebox{0.8}{($0.04$)} &  $0.30$ \scalebox{0.8}{($0.04$)} &  $0.11$ \scalebox{0.8}{($0.04$)} &  $0.26$ \scalebox{0.8}{($0.01$)} &  $0.62$ \scalebox{0.8}{($0.22$)} &  $0.21$ \scalebox{0.8}{($0.01$)} &  $0.65$ \scalebox{0.8}{($0.01$)} &  $0.03$ \scalebox{0.8}{($0.01$)} \\
sghmc gp 1      &  $0.36$ \scalebox{0.8}{($0.08$)} &  $0.38$ \scalebox{0.8}{($0.03$)} &  $0.13$ \scalebox{0.8}{($0.01$)} &  $0.34$ \scalebox{0.8}{($0.01$)} &  $0.15$ \scalebox{0.8}{($0.13$)} &  $0.23$ \scalebox{0.8}{($0.00$)} &  $0.75$ \scalebox{0.8}{($0.01$)} &  $0.04$ \scalebox{0.8}{($0.01$)} \\
sghmc gp 2      &  $0.37$ \scalebox{0.8}{($0.07$)} &  $0.35$ \scalebox{0.8}{($0.03$)} &  $0.13$ \scalebox{0.8}{($0.01$)} &  $0.26$ \scalebox{0.8}{($0.01$)} &  $0.02$ \scalebox{0.8}{($0.01$)} &  $0.23$ \scalebox{0.8}{($0.00$)} &  $0.72$ \scalebox{0.8}{($0.01$)} &  $0.03$ \scalebox{0.8}{($0.01$)} \\
sghmc gp 3      &  $0.38$ \scalebox{0.8}{($0.08$)} &  $0.33$ \scalebox{0.8}{($0.03$)} &  $0.13$ \scalebox{0.8}{($0.01$)} &  $0.25$ \scalebox{0.8}{($0.01$)} &  $0.02$ \scalebox{0.8}{($0.00$)} &  $0.22$ \scalebox{0.8}{($0.00$)} &  $0.71$ \scalebox{0.8}{($0.01$)} &  $0.03$ \scalebox{0.8}{($0.01$)} \\
sghmc gp 4      &  $0.35$ \scalebox{0.8}{($0.09$)} &  $0.31$ \scalebox{0.8}{($0.02$)} &  $0.09$ \scalebox{0.8}{($0.04$)} &  $0.25$ \scalebox{0.8}{($0.01$)} &  $0.02$ \scalebox{0.8}{($0.00$)} &  $0.22$ \scalebox{0.8}{($0.00$)} &  $0.70$ \scalebox{0.8}{($0.01$)} &  $0.02$ \scalebox{0.8}{($0.01$)} \\
sghmc gp 5      &  $0.39$ \scalebox{0.8}{($0.07$)} &  $0.31$ \scalebox{0.8}{($0.02$)} &  $0.13$ \scalebox{0.8}{($0.01$)} &  $0.24$ \scalebox{0.8}{($0.01$)} &  $0.02$ \scalebox{0.8}{($0.00$)} &  $0.22$ \scalebox{0.8}{($0.00$)} &  $0.69$ \scalebox{0.8}{($0.00$)} &  $0.03$ \scalebox{0.8}{($0.01$)} \\
svgp 1          &  $0.33$ \scalebox{0.8}{($0.05$)} &  $0.35$ \scalebox{0.8}{($0.02$)} &  $0.14$ \scalebox{0.8}{($0.01$)} &  $0.32$ \scalebox{0.8}{($0.00$)} &  $0.03$ \scalebox{0.8}{($0.01$)} &  $0.23$ \scalebox{0.8}{($0.00$)} &  $0.73$ \scalebox{0.8}{($0.00$)} &  $0.04$ \scalebox{0.8}{($0.01$)} \\
\bottomrule
\end{tabular}

\end{table}

\end{document}